\documentclass{ieeeaccess}

\usepackage{cite}
\usepackage{graphicx}
\usepackage{array}
\usepackage{booktabs}
\usepackage{caption}
\usepackage{subcaption}
\usepackage{tabularx}
\usepackage{tabularray}
 \usepackage{adjustbox}
\usepackage{amssymb,amsfonts}
\usepackage{textcomp}
\ifCLASSINFOpdf
\else
\fi
\usepackage[cmex10]{amsmath}
\def\BibTeX{{\rm B\kern-.05em{\sc i\kern-.025em b}\kern-.08em
    T\kern-.1667em\lower.7ex\hbox{E}\kern-.125emX}}
\begin{document}
\history{Date of publication xxxx 00, 0000, date of current version xxxx 00, 0000.}
\doi{10.1109/ACCESS.2023.0322000}


\bibliographystyle{IEEEtran}
\title{Detection and Analysis of Stress-Related Posts in Reddit's Acamedic Communities}

\author{\uppercase{Nazzere Oryngozha}\authorrefmark{1}, 
\uppercase{Pakizar Shamoi}\authorrefmark{1}, and 
\uppercase{Ayan Igali}\authorrefmark{1}
}


\address[1]{School of Information Technology and Engineering,
Kazakh-British Technical University,
Almaty, Kazakhstan }


\corresp{Corresponding author: Pakizar Shamoi (e-mail: p.shamoi@kbtu.kz).}

\begin{abstract}
Nowadays, the significance of monitoring stress levels and recognizing early signs of mental illness cannot be overstated. Automatic stress detection in text can proactively help manage stress and protect mental well-being. In today's digital era, social media platforms reflect various communities' psychological well-being and stress levels. This study focuses on detecting and analyzing stress-related posts in Reddit's academic communities.  Due to online education and remote work, these communities have become central for academic discussions and support.
We classify text as stressed or not using natural language processing and machine learning classifiers, with Dreaddit as our training dataset containing labeled Reddit data. Next, we collect and analyze posts from various academic subreddits. 
We identified that the most effective individual feature for stress detection is the Bag of Words, paired with the Logistic Regression classifier, achieving a 77.78\% accuracy rate and an F1 score of 0.79 on the pre-labeled DReaddit dataset. To validate our model's applicability to detect stress in the specific context of academia, we conducted a supplementary experiment by manually annotating 100 posts from academic subreddits, achieving a 72\% accuracy rate. 

Our key findings reveal that the overall stress level in academic texts is 29\%. Posts and comments in professors' Reddit communities are the most stressful compared to other academic levels, including bachelor's, graduate's, and Ph.D. students. This research contributes to our understanding of the stress levels within academic communities. It can help academic institutions and online communities effectively develop measures and interventions to address this issue.
\end{abstract}


\titlepgskip=-21pt

\maketitle


%






\section{Introduction}



%

In our fast-paced world, stress affects countless individuals, impacting their mental well-being.  
Among 48\% of Gen Z adults report feeling anxious, sad, depressed, experiencing FOMO (fear of missing out), and having lowered self-esteem or insecurity\cite{fomo}\cite{fomo2}\cite{gen_z}. According to these findings, Gen Z adults' stress levels may rise due to social media-related worries, highlighting the importance of efficient stress detection and support mechanisms in digital platforms.

Stress statistics worldwide indicate the need for effective stress management \cite{stressstat}. Stress permeates various aspects of people's lives, from the workplace to educational institutions and domestic environments.  Automatic stress detection has the potential to address this global health concern, providing support and resources to those in need. Kazakhstan, like many regions around the world, grapples with the multifaceted challenges of mental health and stress. In the context of this Central Asian nation, several factors have been identified as significant contributors to mental health problems, such as education, traditional belief systems, stigma, and shame\cite{akhand2019factors}.

One key aspect to consider is that individuals may be unaware of their stress levels or underestimate their symptoms. Stress can often go unnoticed until it reaches a critical stage. Automatic textual stress detection can help bridge this awareness gap by identifying stress markers that might go unnoticed.

Moreover, societal factors like the stigma surrounding mental health and the hesitance to seek professional help create barriers to receiving timely support \cite{stigma1} \cite{stigma2}.  While psychological help is valuable, it can be expensive and inaccessible for many individuals. Automatic stress detection can be an accessible and complementary tool to traditional services, assessing stress levels without direct interaction with specialists. This tool can help individuals overcome hesitations and encourage early intervention by specialists by identifying individuals who require further assistance.

As for the existing psychological tests, they rely heavily on self-reporting \cite{selfreport}, which can be influenced by biases or individuals not acknowledging their stress levels. Automatic stress detection of texts in social media offers a data-driven approach, reducing reliance on self-reported data.


In the dynamic landscape of higher education, marked by rigorous academic demands and the pursuit of future goals, stress has become a pervasive issue for both students and pedagogical staff \cite{Hong_Kong, Malaysian, stressFactors_nbci}. 
The other study reveals that students aged 18-19 exhibited more symptoms of common mental disorders than non-students, though the effect size was small\cite{McCloud2023}. Another study identified that a substantial portion of Ph.D. participants were at high risk for depression \cite{Friedrich2023}. It was found that unhealthy working conditions and increased mental health risks among Ph.D. students. Identified life satisfaction, perceived stress, and negative support as primary predictors of anxiety and depression, and the impact of supervision and the work environment on the mental health and overall well-being of Ph.D. students. Detecting and addressing stress early is essential to prevent it from developing into more severe issues like anxiety or burnout.

Nowadays, people share their stressful experiences and challenges with mental health through online forums, micro-blogs, and tweets as Internet usage has grown \cite{users}. A 2019 survey conducted by Hill Holiday revealed that 94\% of Gen Z adults (individuals born between the mid-1990s and early 2010s) reported having at least one social media account\cite{gen_z}.
So, the alarming prevalence of stress worldwide highlights the urgent need for innovative approaches to address this issue.

Reddit is one of the most popular social networks. It consists of thousands of communities called ``subreddits'' (r/subredditname) that focus on specific interests, hobbies, or themes. Every post strongly follows the topic and rules of the subreddit where it was posted, which makes it easy to analyze the behavior and opinions of the community. Posts on Reddit have a score that is responsible for popularity and it is calculated with the sum of upvotes and downvotes. Reddit was chosen as the data source due to its community's open and supportive nature, allowing longer posts where people readily share their problems, including those related to anxiety \cite{stress-dep2}. This choice aligns with the importance of studying online platforms like Reddit in the context of well-being and the recognition of the presence of toxicity and anxiety within Reddit communities \cite{toxic2} \cite{toxic}.



\begin{figure}[t]
\centerline{\includegraphics[width=0.48\textwidth]{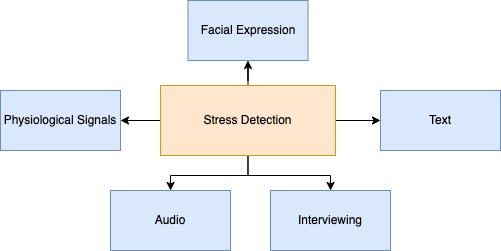}}
\caption{Main techniques to detect stress.}
\label{waystodetect}
\end{figure}


The present paper aims to use Natural Language Processing (NLP) and Machine Learning (ML) classification algorithms to automatically detect stress in social media such as Reddit.  We further apply the proposed method to analyze the stress levels in texts of professors, bachelor's, graduate, and Ph.D. students. For that, we collected and analyzed a dataset containing 1,584 posts and 122,684 comments on academia-related topics from the Reddit social network.


The contributions of this paper are presented as follows:
\begin{itemize}
\item Comparative analysis of ML classifiers for stress detection
    \item Evaluation of the applicability of the method in the context of academia. We initiated the annotation process to evaluate our method further due to the absence of available academic stress-related datasets labeled by humans. 
    \item Identification of the prevailing emotions associated with stress messages at different levels of academia.
    \item Understanding stress patterns across various academic levels, including professors, Ph.D. students, graduate students, and bachelor students.
    \item Use of long-length text that can capture stress indicators.

\end{itemize}

The rest of the paper is organized as follows. Section I is this Introduction. Section II provides a literature review of research in stress detection. Section III describes Methods, including data preprocessing, feature extraction, and classification. Data collection, description, and annotation procedures are also presented there. Section IV compares the results and presents the most powerful machine-learning method for stress detection. We also present the results of stress evaluation within Reddit's academic communities. Section V compares our results with those of similar studies. Finally, Section VI offers concluding remarks. Additionally, it offers suggestions for how the technique might be enhanced in the future.

The following section provides an overview of stress detection techniques.

\section{Related work}

This section provides an overview of stress detection studies in Social Media. 

There are different ways of detecting stress, including Physiological Signals \cite{Chauhan2017, review1, wearbleSensors} (e.g., Heart Rate, Humidity response, Temperature response), Audio \cite{audio_stress, audio_stress_2}, Facial Expression \cite{Raichur2017}, Interviewing or Psychological tests \cite{PSS_test, PSLE_test, DHUS_test}, and Text-based approaches \cite{Nijhawan2022, stress3, Turcan2019, munoz} (see Figure \ref{waystodetect}). 

Many text-based approaches for stress detection were proposed, and some of them identified that the Bidirectional Long Short-Term Memory (BLSTM) with attention mechanism is the most effective approach \cite{stress1}, \cite{stress6}. The other study focuses on continuous lexical embedding and the BLSTM model \cite{stress4}.  Another study introduces the KC-Net, a mental state knowledge–aware and contrastive network effective for early stress detection \cite{stress5}.  A light and robust method to detect depressive texts in social media was proposed in \cite{stress7}. The authors used an attention-based bidirectional LSTM and CNN-based model. Some works focus on applying the SVM-based approach \cite{stress8}, achieving the 0.75 accuracy. To increase classification accuracy, some approaches combine lexicon-based features with
distributional representations \cite{stress9}. The other work proposes a TensiStrength method to identify the stress or relaxation level in the text, it applies a lexical approach and a set of rules \cite{stress10}. 


Some works concentrate on the analysis of textual data in social media. In the study of Nijhawan et al. \cite{Nijhawan2022}, the authors employ large Twitter datasets, machine learning techniques, and BERT to classify sentiment. They also use Latent Dirichlet Allocation, an unsupervised ML approach, to find document patterns and forecast pertinent subjects. These algorithms simplify the process of identifying people's feelings on internet platforms.

 In the other study, authors look at the use of emotion detection for explainable psychological stress detection \cite{Turcan2}. They research multi-task learning and the empathetic modification of language models. Their models, trained by the Dreaddit dataset \cite{Turcan2019}, produce results comparable to BERT models. Another work that used the dataset Dreaddit is \cite{Inamdar2023}. They used various NLP tools, BERT tokenizers, and a Bag of Words method to prepare textual data for machine learning models. Each method's outcomes are reviewed, with an accuracy of 76\%.   The study  \cite{jadhav2019} classified the Dreaddit data using the VM, LSTM, and BLSTM algorithms, and reported that BLSTM had the best accuracy. However, they did not specify how much accuracy was achieved.

Authors of the related paper \cite{munoz} discuss the suggested technique for classifying stress, which combines Word2Vec, GloVe, and FastText word embeddings with lexicon-based characteristics. It gets strong classification results with 84.87\% F-Score utilizing word embeddings on the Dreaddit dataset and 82.57\% F-Score using lexicon-based features. While balancing performance and computing resources, their approach shows potential in reliably diagnosing stress from the text. Notably, the technique beats more intricate systems like BERT.

Next, the other study presented Stress Annotated Dataset (SAD)\cite{Mauriello2021}, and by developing a model and exploring how it may be used in mental health contexts, they demonstrate how the dataset can be used in real-world situations. The study provides a classification method for everyday stress, a collection of SMS-like phrases that represent stressors, and an assessment of the method's efficiency in categorizing various themes.

While Chauhan et al. \cite{Chauhan2017}, \cite{review1} concentrate on physiological markers like body temperature, GSR, and blood pressure to detect stress, Raichur et al. \cite{Raichur2017} use image-processing techniques. Both strategies, however, have limitations since they depend on actual presence. While physiological measures demand specialized equipment and can be intrusive, image processing requires controlled conditions and may not be effective in all circumstances. Researchers are investigating several non-invasive, scalable approaches, such as analyzing textual data, voice, and behavioral clues to solve these constraints. Researchers want to create more adaptable and user-friendly methods for precisely detecting stress levels by extending the spectrum of stress detection techniques beyond a physical presence.


As we can see, stress detection is challenging and worthy of further exploration for several reasons, including subjectivity, context dependency, and multimodal nature.  Research on textual stress detection is still limited in using short-length texts for analysis. Short-length text can be enough for sentiment analysis, but when it comes to detecting mental health problems detections, including stress detection it would be proper to use long texts as they are more informative\cite{ieeeAccess_sysReview}.

\section{Methods}
\begin{figure*}[htbp]
\centerline{\includegraphics[width=\textwidth]{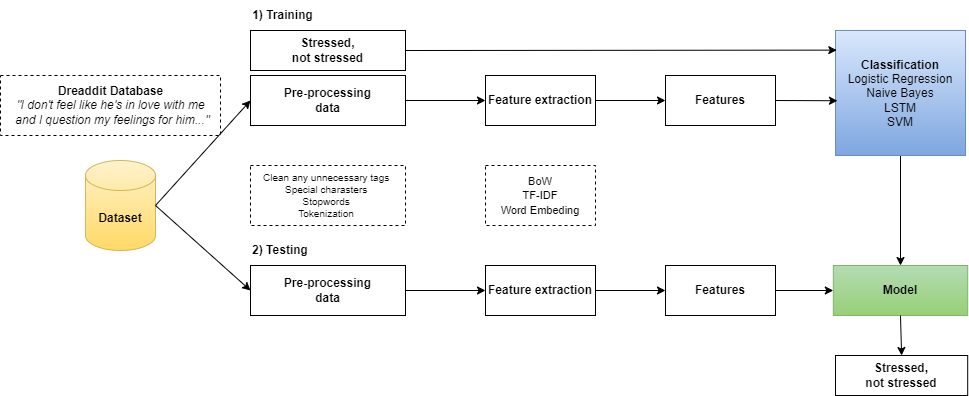}}
\caption{Schematic representation of the proposed approach. As a training dataset, we use the Dreaddit pre-labeled dataset \cite{Turcan2019}.}
\label{method}
\end{figure*}

In this research paper, we build a system that classifies whether the text is stressful. The schematic representation of the approach is shown in the Figure \ref{method}. To implement it, we need to collect data from various sources or datasets and use the NLP techniques such as word embedding, and Bag of Words(BoW). They are intended to create word/sentence data that can be fed to ML models.

\subsection{Data Collection}
We used the following datasets in our study:
\subsubsection{DReaddit}
To train the model, we used Dreaddit, a text data collection specifically designed for detecting stress\cite{Turcan2019}. It is a publicly available dataset that contains a collection of Reddit posts and associated metadata. A total of 190 thousand posts from Reddit are included in the dataset, of which 3553 were labeled manually. The authors offer initially supervised learning algorithms for stress detection that combine neural and traditional methodologies. The data corpus contains stress-indicative posts in 5 domains: abuse (703 posts), anxiety (728 posts), financial (717 posts), PTSD (711 posts), and social (694 posts). Table \ref{domain} demonstrates the examples of stressed posts for each domain.

\begin{table*}[]
\caption{Stressful Dreaddit posts with domains}
\begin{center}
\begin{tabular}{|m{48em}|m{5em}|m{5em}|}
\hline
\centering
            Post &  Stress label &  Domain \\
\hline
         I've mostly come to terms with it but every time I see his name or face or his girlfriend's it comes back. My question is, when the justice system fails, how do you deal with living near a child molester when you can't just pick up and leave?.. & 1 & abuse \\
          & & \\
          \hline
I cried for hours and at one point, something came over me and just slammed my head into my bathroom door. Sadly, since I'm in a dorm, it's a shitty hollow core door and it broke bad and now there is a hole that I have to figure out how to fix. It's a \$100 fine if I can't figure out what to do with it so that's just compounding on the already existing stress. ... & 1 & anxiety \\
& & \\
\hline
I asked for nothing but a declaration (a document detailing finances) from the divorce. He stole from me. I asked for nothing but restitution. He fought, forced me to hire a lawyer (more money), dragged out the case, and led to having alimony imposed upon if he failed to make restitution... & 1 & financial \\
& & \\
\hline
I cried for hours and at one point, something came over me and just slammed my head into my bathroom door. Sadly, since I'm in a dorm, it's a shitty hollow core door and it broke bad and now there is a hole that I have to figure out how to fix. It's a \$100 fine if I can't figure out what to do with it so that's just compounding on the already existing stress. ... & 1 & ptsd \\
& & \\
\hline
That's it. My mom made me delete it and said ``that's what is wrong with your generation, you act too grown''. Nothing about my picture was ``grown'' I was actually wearing a long white t-shirt and boyfriend jeans. Even when we go to the beach with my siblings, me and my sister were wearing the exact same bikini and she said something to me about ``needing to sit my fast-tailed butt down''... & 1 & social \\
& & \\
\hline
\end{tabular}
\end{center}
\label{domain}
\end{table*}



\subsubsection{Test Datasets - Reddit Academic Communities}

For this study, data was collected from the most popular posts based on scores of various subreddits connected with academic topics (``r/Professors'', ``r/PhD'', ``r/GradSchool'', ``r/csMajors'', and ``r/EngineeringStudents'') from Reddit using a Python module ``PRAW'' \footnote{https://github.com/praw-dev/praw}(Python Reddit API Wrapper).

PRAW is a Python library that provides easy access to Reddit's API. PRAW aspires to be user-friendly and adheres to all of Reddit's API restrictions. Reddit API has limits, saying it allows 60 requests for one minute.

\begin{table}[]
\caption{Number of posts and comments from each subreddit}
\centering
\begin{tabular}{@{}|l|l|l|@{}}
\hline
            \textbf{Subreddit} &  \textbf{\# of posts} &  \textbf{\# of comments} \\
\hline
         r/Professors &  313 &   37235 \\
                r/PhD &  386 &   22406 \\
         r/GradSchool &  785 &   44762 \\
           r/csMajors &   70 &    9436 \\
r/EngineeringStudents &   30 &    8845 \\
\hline
                      & 1584 &  122684 \\
\hline
\end{tabular}
\label{distrb}
\end{table}

\subsection{Data Description}

Using Reddit Scrapper we collected 1584 posts and 122684 comments for the posts. The data was collected on June 2, 2023. Each piece of data has information about \textit{Date, Title, Text, Score, Tag.} Reddit is frequently used as a platform for online discussions within various communities or ``subreddits''. Table \ref{distrb} illustrates the distribution of posts and comments among subreddits that were scrapped. We only took those posts with ``self-text,'' the main text beside the title. On Reddit users can make posts with title and image, video and gif, or title and text, it is impossible to combine them. This explains why r/csMajors and r/EngineeringStudents have less data: they tend to post memes more than graduates or professors and usually do not categorize their posts with tags.

Posts on Reddit can have a flair or tag; the number and topic of tags vary from subreddit to subreddit. They specify the post's topic, and by looking at tags, we can understand what kind of posts are usually popular and get attention. Table \ref{prof}, Table \ref{mast}, Table \ref{phd}, and Table \ref{bach} present the distribution of scraped posts and comments across various tags. Even if not every post was tagged, we can see some tendencies. Posting memes are quite common for all communities (``Humor'', ``Fun and \& Humor'', ``Shitposting'', ``Memes''). Another theme common to all communities is asking for advice, which is a general topic on Reddit, as each subreddit can be considered a separate online forum. Professors and Ph.D. students/graduates have the same habit of venting. 

\begin{table}[]
\caption{Professors tags}
\centering
\begin{tabular}{@{}|l|l|l|@{}}
\hline
\textbf{Tag}                           & \textbf{\# of posts} & \textbf{\# of comments}   \\ 
\hline
Rants / Vents                 & 40              & 5968                 \\
Humor                         & 18              & 1892               \\
Advice / Support              & 6               & 1320              \\
Academic Integrity            & 3               & 296                \\
Teaching / Pedagogy           & 3               & 358                \\
Mindblown                     & 1               & 336                \\
COVID-19                      & 1               & 69                 \\
Other (Editable)              & 1               & 28                 \\
Where Humor and Rants Collide & 1               & 159                \\
FML                           & 1               & 83                 \\
Technology                    & 1               & 118                \\
Untagged                      & 237             & 26608              \\
\hline
                              & 313             & 37235             \\ \hline 
\end{tabular}

\label{prof}
\end{table}

\begin{table}[]
\caption{Graduates tags}
\centering
    
\begin{tabular}{@{}|l|l|l|@{}}
\hline
\textbf{Tag}                           & \textbf{\# of posts} & \textbf{\# of comments}   \\ \hline
Health \& Work/Life Balance & 84              & 5908               \\
Fun \& Humour               & 36              & 2185               \\
Academics                   & 38              & 1804               \\
Research                    & 18              & 998               \\
Professional                & 8               & 639                \\
Finance                     & 8               & 796                 \\
Admissions \& Applications  & 9               & 589                \\
News                        & 4               & 295                \\
Untagged                    & 580             & 31548              \\ \hline
                            & 785             & 44762              \\ \hline
\end{tabular}
\label{mast}
\end{table}

\begin{table}[]
\caption{PhD tags}
\centering
\begin{tabular}{@{}|l|l|l|@{}}
\hline
\textbf{Tag}                           & \textbf{\# of posts} & \textbf{\# of comments}
\\ \hline
Vent             & 128             & 9684               \\
Post-PhD         & 36              & 1610               \\
Dissertation     & 75              & 3197               \\
Other            & 92              & 4975               \\
Humor            & 26              & 1490               \\
Admissions       & 3               & 134                \\
Preliminary Exam & 7               & 260                \\
Need Advice      & 10              & 773                \\
Untagged         & 9               & 283                \\ \hline
                 & 386             & 22406              \\ \hline
\end{tabular}

\label{phd}
\end{table}

\begin{table}[]
\caption{Bachelors tags (csMajors and EngineeringStudents)}

\centering
\begin{tabular}{@{}|l|l|l|@{}}
\hline
\textbf{Tag}                           & \textbf{\# of posts} & \textbf{\# of comments}   \\ \hline
Shitpost         & 12              & 818                \\
Flex             & 4               & 296                \\
Others           & 4               & 770                \\
Rant             & 5               & 972                \\
Company Question & 2               & 91                 \\
Advice           & 3               & 541                 \\
Course Help & 1               & 303                
\\
Untagged         & 69              & 14490               \\ \hline
                 & 100              & 18281               \\ \hline
\end{tabular}
\label{bach}
\end{table}

\begin{figure*}[tb]
    \begin{subfigure}{0.25\textwidth}
        \includegraphics[width=\textwidth]{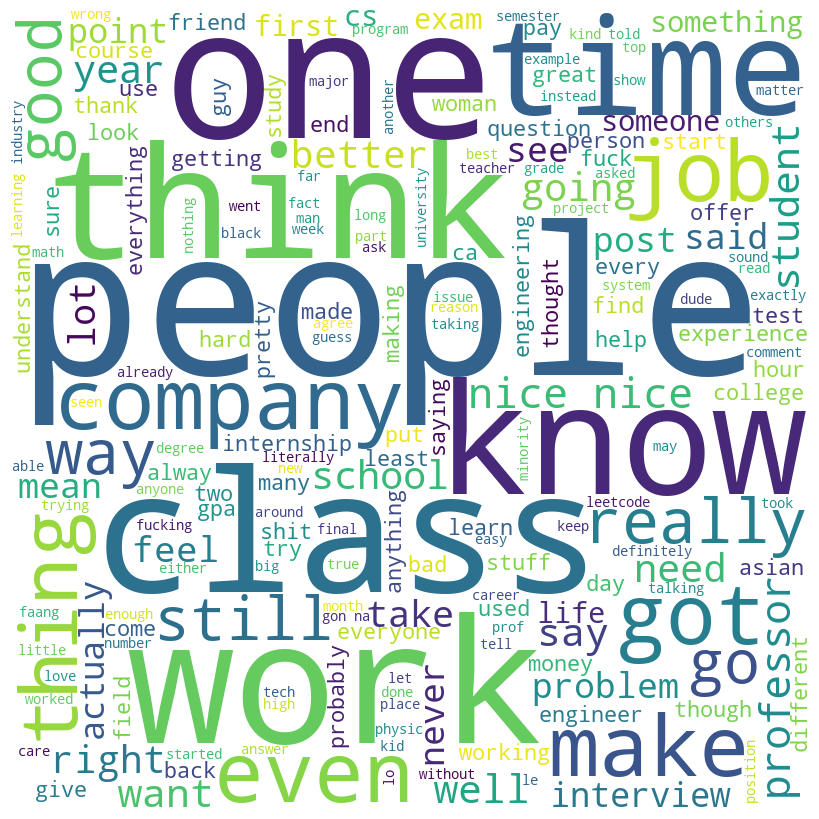}
        \caption{Bachelor students}
    \end{subfigure}%
    \hfill
    \begin{subfigure}{0.25\textwidth}
        \includegraphics[width=\textwidth]{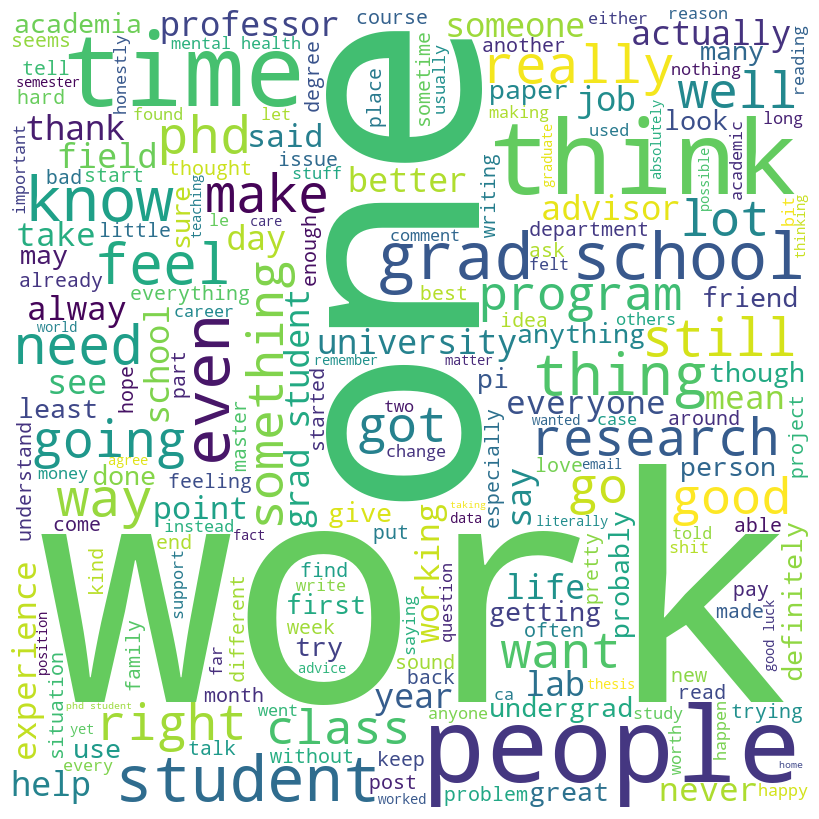}
        \caption{Graduate students}
    \end{subfigure}%
    \hfill
    \begin{subfigure}{0.25\textwidth}
        \includegraphics[width=\textwidth]{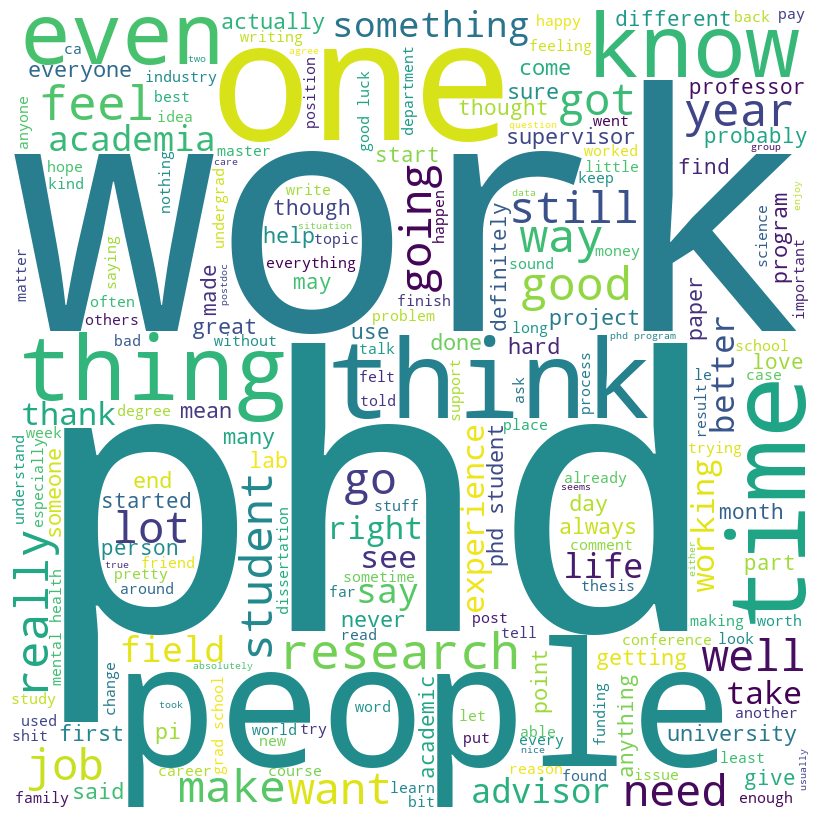}
        \caption{Ph.D. students}
    \end{subfigure}%
    \hfill
    \begin{subfigure}{0.25\textwidth}     
        \includegraphics[width=\textwidth]{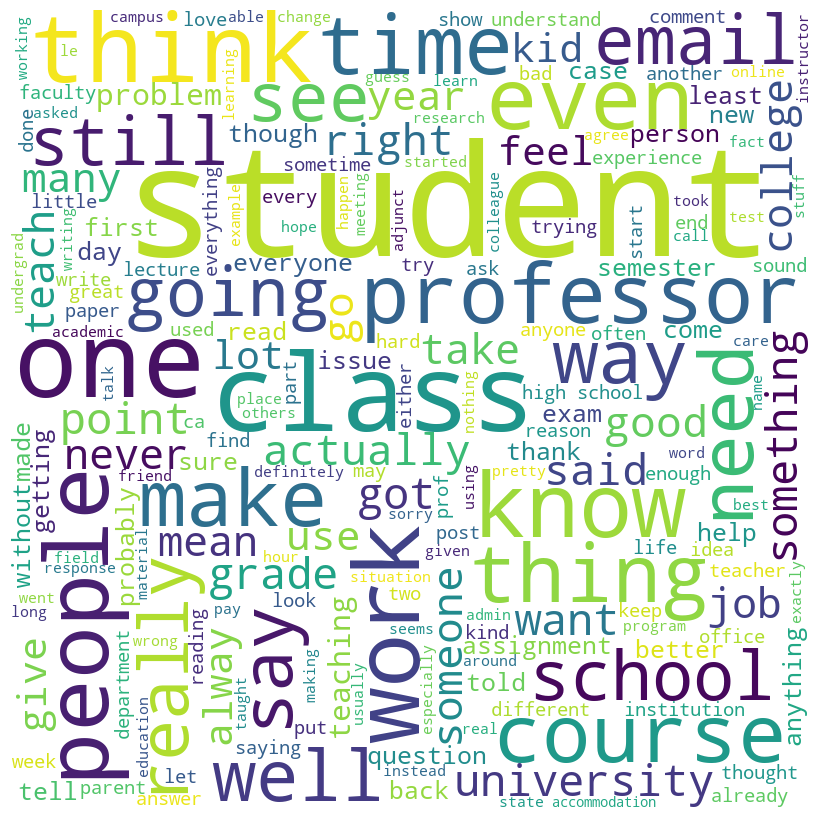}
        \caption{Professors}
    \end{subfigure}
    \caption{Word clouds.}
    \label{subreddits_wordclouds}
\end{figure*}

Next, Figure \ref{subreddits_wordclouds} presents word clouds of text from posts we scrapped from Reddit communities. Common words for all communities are ``think''(13330 overall occurrences from all communities), ``time'' (19460), ``work'' (18730) and ``people'' (16842). In total, the Reddit data encompassed 43,209 distinct words. Figure \ref{subreddits_wordclouds} also shows us that bachelors write a lot about getting a job, companies, interviews, and internships. This can be because our data is from Computer Science and Engineering communities. Bachelors used 16085 unique words. 

Graduates and Ph.D. students are different with words like ``dissertation'', ``advisor'', ``research'', ``paper'', ``lab'', ``academic'' or ``academia'', etc. Graduate and Ph.D. students used 32449 unique words. In addition, Figure \ref{subreddits_wordclouds} illustrates that professors mainly discuss students, emails, classes, teaching, grading/evaluation, parents, and online education. Professors used 25266 unique words. Bachelors and professors have a connection regarding words like ``lecture'', ``class'', ``professor'', and ``grade''.

\subsection{Stress Detection}
\subsubsection{Pre-processing}
\begin{figure*}[tb]
\centerline{\includegraphics[width=\textwidth]{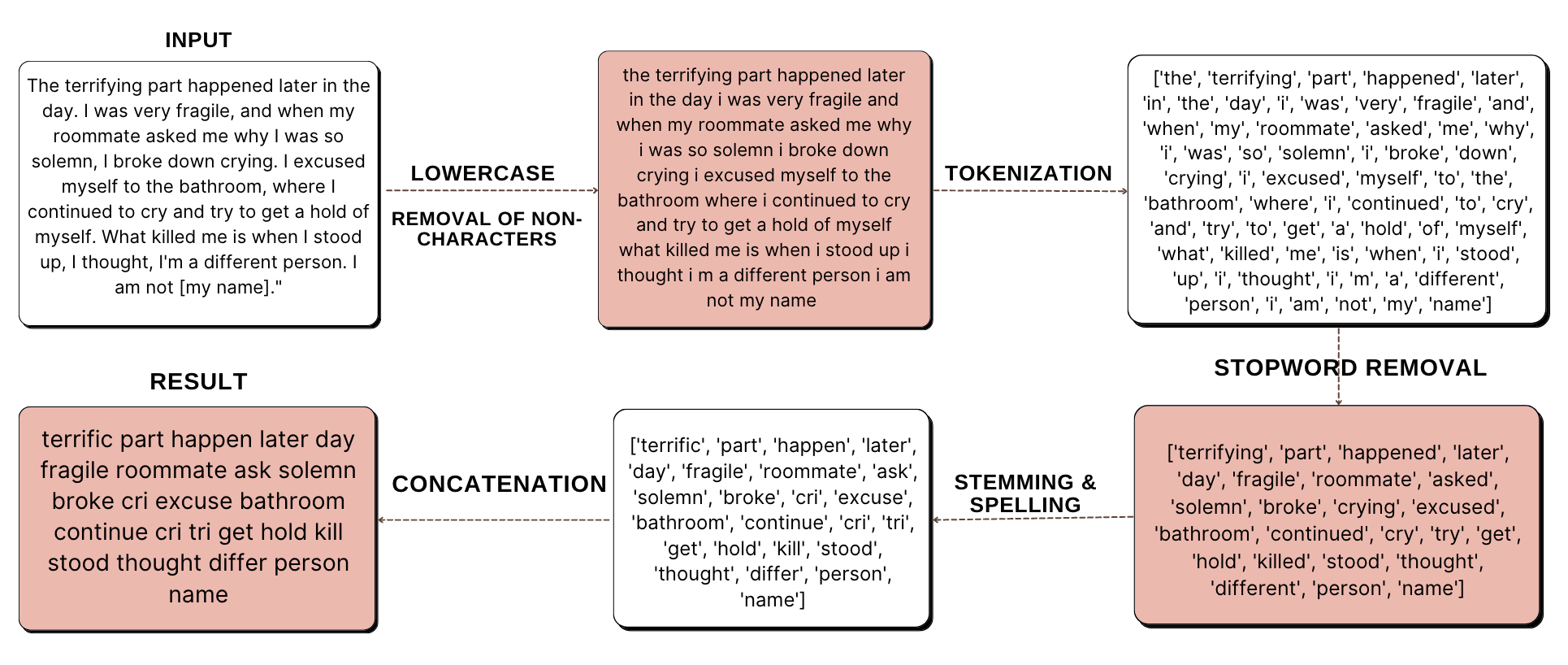}}
\caption{
Illustration of preprocessing steps.}
\label{pre_processing_steps}
\end{figure*}

\begin{figure}[tb]
\centerline{\includegraphics[width=9cm,scale=0.3]{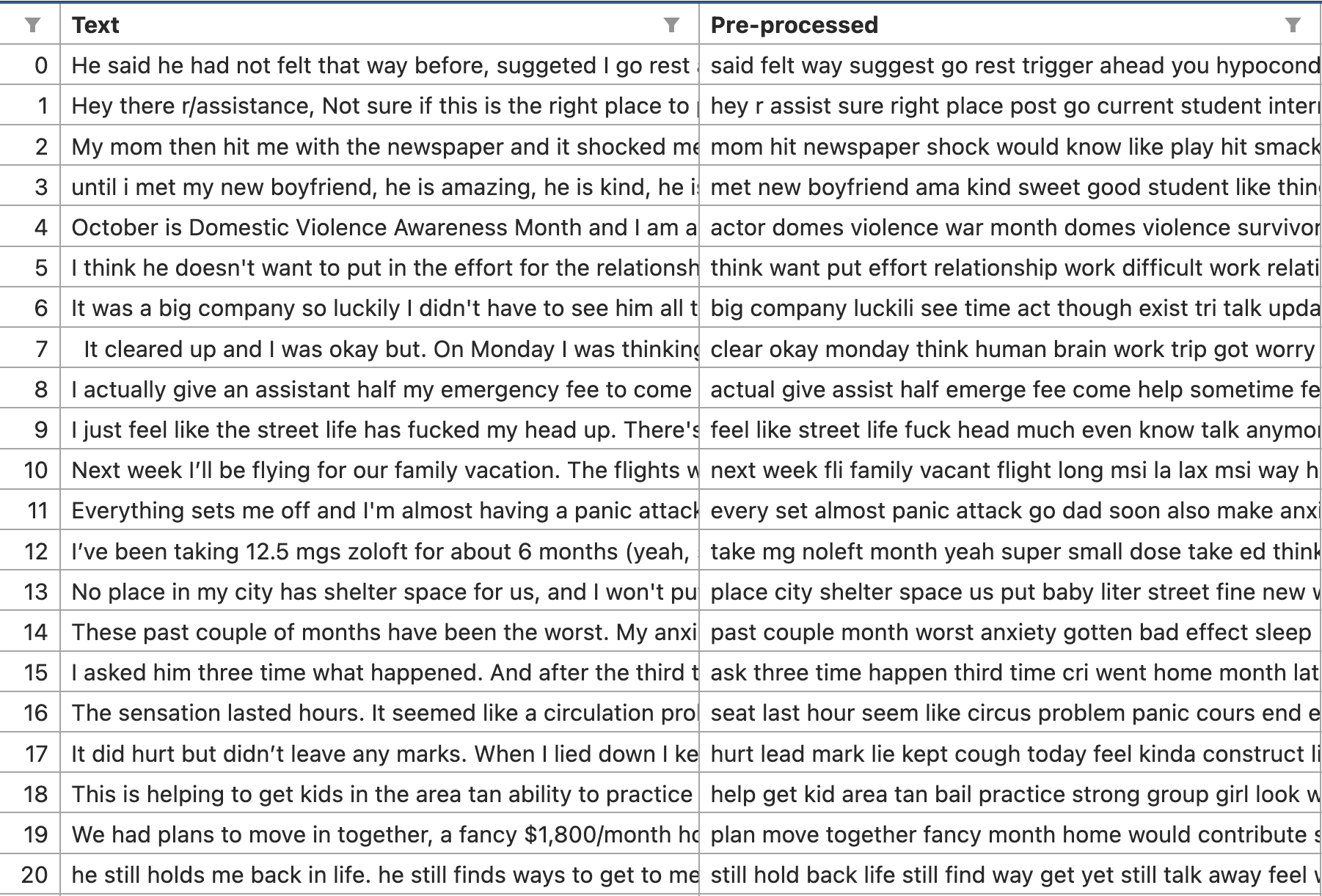}}
\caption{After pre-processing.}
\label{pre_proc}
\end{figure}

Before applying ML algorithms, posts were preprocessed. During the preprocessing stage, various necessary adjustments were made to prepare the text data for ML classification. The preprocessing steps applied to the dataset are as follows (see Figure \ref{pre_processing_steps}):
\begin{enumerate}
\item \textbf{Lowercase:} All text data has been converted to lowercase to ensure consistency and uniformity of references.
\item \textbf{Removal of non-characters:} Removed non-characters, such as letters and special characters, from the text data: removing HTML tags, ``@'', ``\_'', and other special characters. This step helps eliminate unnecessary noise and information. 
\item \textbf{Tokenization:} The text was tokenized using a tokenizer, which splits the text into individual tokens. This step is necessary to divide the text into basic units.
\item \textbf{Stopword Removal:} Common English stopwords (e.g., ``and'', ``the'', ``in'') are removed from the tokenized text. Grammatical terms are often discarded because they have no special meaning for classification tasks.
\item \textbf{Stemming:} The stem process was used to reduce the remaining words to their original form. This helps form similar words and simplifies vocabulary.
\item \textbf{Combining Words:} Finally, the preprocessed words are recombined as coherent text strings, ready to be used in machine learning classification.
\end{enumerate}
Together, these preprocessing steps improve the quality and usability of the texture data, making it more suitable for subsequent ML classification tasks. The difference between raw and processed data can be seen in Fig. \ref{pre_proc}.


If the data is labeled, it goes to the classification step; otherwise, it goes to the Feature extraction step.

\subsubsection{Feature extraction}
Reducing the dimensionality of the original raw data is achieved by feature extraction, resulting in more manageable groupings that can be processed quickly \cite{feature_exp}. Different feature extraction techniques, like BERT embeddings, Word2Vec with TF-IDF weights, bigram TF-IDF, unigram TF-IDF, and Bag of Words, are available. In this work, we use Bag of Words.

\paragraph{Bag of Words}

\begin{figure*}[tb]
    \centering
    \includegraphics[width=\textwidth]{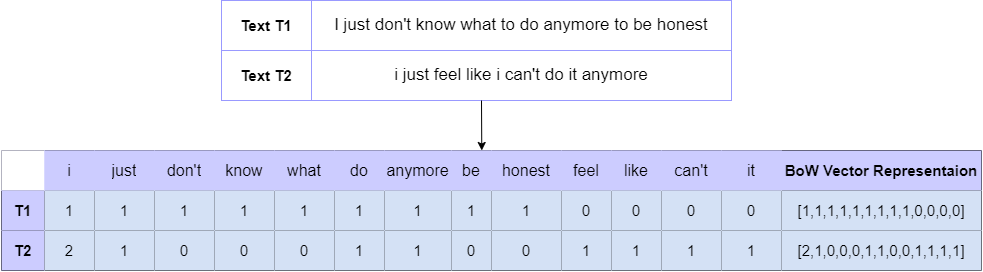}
    \caption{Bag-of-Words work principle.}
    \label{bow_fig}
\end{figure*}

The principle of BoW is computing the frequency of occurrence for each word while considering the complete text data \cite{feature_exp}. Figure \ref{bow_fig} illustrates the principle of how Bag-of-Words works. Following that, the terms are mapped to their corresponding frequencies. This method ignores word order while accounting for the frequency of each word.

\paragraph{TF-IDF}
A statistical tool called TF-IDF evaluates a word's significance inside a document or set of documents. It combines inverse document frequency (IDF), representing a word's rarity throughout the corpus, with term frequency (TF), which determines how frequently a word appears in a text. To capture their relative significance, TF-IDF gives terms common in a text but uncommon in the corpus larger weights.

\paragraph{Word Embedding}
A feature of the learning method called word embedding uses the contextual hierarchy of words to map them to vectors. The feature vectors of comparable words will be the same. For example, the places of the words ``dog'' and ``puppy'' are closer, because of their meaning.

\subsubsection{Classification}

To estimate the presence of stress in text, we employ approaches to classification to estimate the likelihood of stress within the users. The following ML algorithms have been used:   
\paragraph{SVM}
An ML approach called the Support Vector Machine (SVM) is generally employed for classification tasks. It establishes a decision boundary that divides the data into several classes to optimize the margin or separation between the decision boundary and the support vectors. Using kernel functions, SVM can handle data that may be separated into linear and non-linear categories. It has been widely used for successful categorization across many fields \cite{ieeeAccess_sysReview, wearbleSensors, collegeStuds_SVA, multimodalPhysData}.

\paragraph{Logistic Regression}
 Consider a contextual setting within regression analysis wherein the dependent variable exclusively manifests two potential outcomes, specifically 0 or 1. This configuration aligns with the framework of binary logistic regression. In contrast, applying multinomial logistic regression becomes pertinent when the dependent variable encompasses a multitude of potential outcomes that surpass the binary domain. These values might be assigned based on qualitative observations. Consider a model structured as follows: $ y_i =  x'_i\beta + \varepsilon_i \quad x'_i = [1, x_{i1}, x_{i2},.., x_{ik}]$ \cite{linear_book}. Because of its interpretability and simplicity, it is a well-known algorithm that can quickly train and evaluate \cite{ieeeAccess_sysReview, wearbleSensors}.

The Binary Logistic Regression model estimates the probability that the dependent variable \(Y\) equals 1 given input \(X\). It uses the logistic (sigmoid) function to transform a linear combination of input features into a probability value between 0 and 1 as in Equation \ref{eq: logistic}. 

\begin{equation}
\label{eq: logistic}
P(Y=1|X) = \frac{1}{1 + e^{-(\beta_0 + \beta_1X_1 + \beta_2X_2 + \ldots + \beta_nX_n)}}
\end{equation}

Where:
$P(Y=1|X)$ is the likelihood that $Y$ equals 1 given input $X$. The model coefficients $\beta_0, \beta_1, \beta_2, \ldots, \beta_n$ are values learned during training, and $X_1, X_2, \ldots, X_n$ represent the input features.


The logistic function converts a linear combination of input data into a probability value between 0 and 1. The data point is given to the positive class (1) if the estimated probability is greater than or equal to 0.5; otherwise, it is assigned to the negative class (0).

\paragraph{Naive Bayes}
An efficient probabilistic approach for classification tasks is Naive Bayes. Given the class label, it assumes that the characteristics are conditionally independent, making the probability calculation easier. The effectiveness and scalability of Naive Bayes in handling huge datasets are well established\cite{ieeeAccess_sysReview, wearbleSensors, collegeStuds_SVA}.

\paragraph{LSTM}
Long Short-Term Memory (LSTM) can recognize long-term relationships in sequential data. It is commonly utilized in jobs involving sequential data and natural language processing. To effectively simulate sequential patterns, LSTM features an internal memory system that enables it to preserve and selectively forget information across extended durations \cite{lstm1,lstm2}.


\subsection{Dataset Annotation on an 11-point scale}
We initiated the annotation process to evaluate our method due to the absence of available academic stress-related datasets labeled by humans. The trial dataset comprised 100 posts or comments (25 for each academic level) randomly sampled from our unlabeled stress-related Reddit dataset. 5 annotators, including four regular humans and one psychologist, assessed the stress levels of the posts using an 11-point scale. Each annotator was instructed to score the sentiment from -5 (strong stress) to +5 (unstressed). 

\begin{figure*}[t]
\centering
\includegraphics[width=\textwidth]{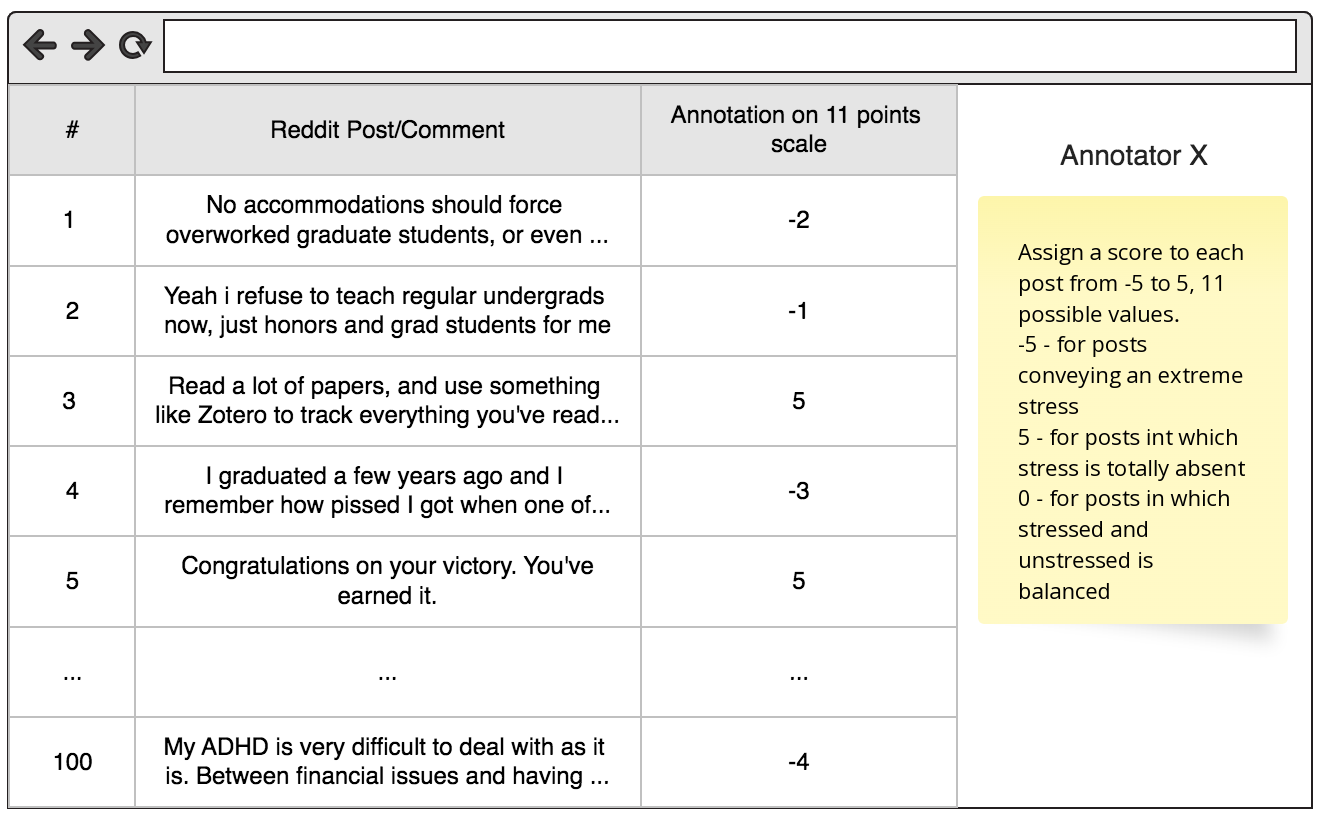}
\caption{Sample form (Google Sheets) with Reddit posts or comments for annotation provided to each annotator. Score descriptions and data annotation methods were taken from the following study \cite{rosso}. The page design was taken from \cite{peerj}. 
The form contains 100 posts (25 posts each in 4 categories) from our unlabeled stress dataset.
}
\label{annform}
\end{figure*}

It is desired that data annotation is done by someone who has a full understanding of the topic, hence a psychologist (Ph.D. in Psychology) was requested to contribute. She was given double weight because she is more familiar with the context's terminology. The human raters labeled the stress levels of the posts using separate Google sheets obtained online. Figure~\ref{annform} shows a screenshot of the form (Google Sheets).


We applied the strategies described in \cite{rosso}: if 60\% or more of annotator labels are considered outliers, the annotator judgments are removed from the job. However, we decreased the threshold to 40\% for more accurate resultant annotations.

We use the following formula - Equation \ref{eq:score} to identify if a judgement $A_{i,j}$ is an outlier \cite{rosso}:
\begin{equation}
\label{eq:score}
    |A_{i,j}-avg(A_{i',j})| > std_{t}(t_{j}),
\end{equation}
where $i$ is the index of a particular annotator, and $j$ is the index of a specific tweet/post. The notation $\text std_{t}(t_{j})$ is the standard deviation of all scores given for a tweet/post $t_{j}$.

One annotator, with a 41\% outlier rate, had their annotations excluded from the analysis, and we recalculated the weighted average scores for each post.



\section{Experimental Results}

\subsection{Accuracy evaluation}
\begin{figure*}
    \includegraphics[width=\textwidth]{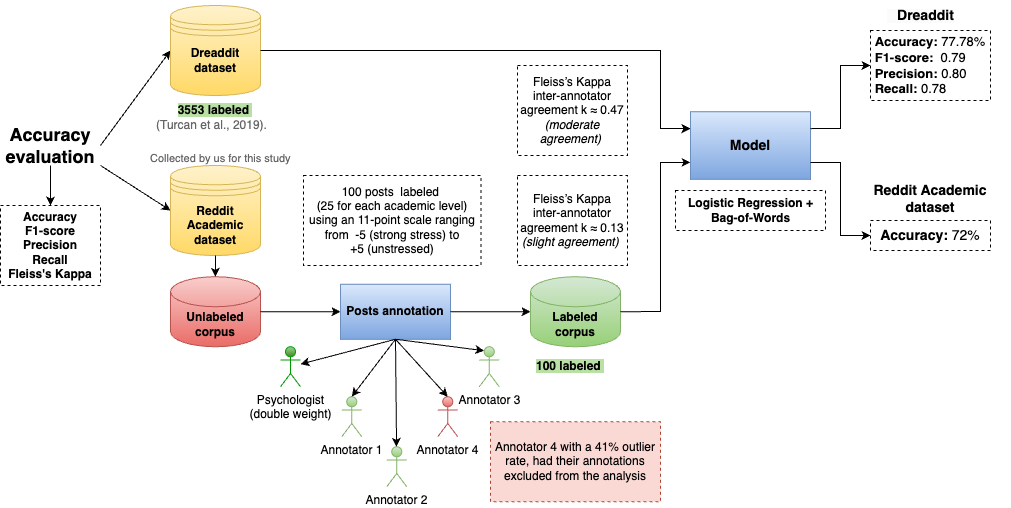}
    \caption{Representation of the accuracy evaluation on the Dreaddit dataset \cite{Turcan2019} and on the Reddit Academic Dataset.}
    \label{accuracy_eval}
\end{figure*}

Figure \ref{accuracy_eval} illustrates a schematic representation of how the accuracy of the proposed model was evaluated in our study using different datasets.
In the initial evaluation phase, the Dreaddit dataset, comprising 3553 human-labeled posts, served as the foundational training set for our model. The accuracy of the model reached 77.78\% on this dataset when employing Logistic Regression and the Bag of Words methodology.

Following this, in the academic evaluative phase, we employed a collection of unlabeled posts from an academic context. We performed data annotation on a small dataset involving five annotators (including a psychologist) and applied the model, resulting in an accuracy of 72\%. This comprehensive approach allowed us to evaluate the model's performance thoroughly. The detailed description is provided in the following subsections.

\subsubsection{Dreaddit (Training Dataset)}
The feature extraction method BoW was paired with the SVM, Naive Bayes, Logistic regression, and Word Embeddings with the LSTM algorithm to classify stressed or non-stressed texts. 

Table \ref{example} shows the result of a preprocessing tool, in cases when the model made incorrect predictions because of preprocessing. The text must be classified as stressed in the first row, but let us look at the preprocessed data and consider the words separately. If we look at the words ``mom'', ``hit'', ``newspaper'', ``shock'', ``like'', ``play'', it is difficult to understand that in the original text, the mother hits the author with the newspaper, and that's why the author was shocked. A recap of the third text is that the author helps the homeless lady. However, the words of preprocessed data made the model think that the author may be an anxious person or a homeless who needs food and money.

\begin{table}[]
\caption{The output of prediction of the model BoW + Logistic Regression}
\begin{tabular}{|m{3cm}| m{2cm} |m{1cm} |m{1cm}|}
\hline
\centering
\textbf{Text} & \textbf{Preprocessed} & \textbf{Expected} & \textbf{Actual}   \\

\hline
        My mom then hit me with the newspaper and it shocked me that she would do this, she knows I don't like play hitting, smacking, striking, hitting or violence of any sort on my person... & mom hit newspaper shock would know like play hit smack strike hit violence sort person ...send vibe ask universe yesterday decide take friend go help not friend move new place drive friend move strike shoulder & 1 & 0  \\ 
        & & & \\
        \hline
        No place in my city has shelter space for us, and I won't put my baby on the literal street... & place city shelter space us put baby liter street ... & 1 & 1 \\ 
        & & & \\
        \hline
        but I'm really, really afraid of public embarrassment and awkward situations. So I was in the train station and saw this homeless lady asking for food and money. I always help homeless people if I have a change in my wallet. I walked up to her, took out my wallet, and pulled out this £5 cash to give her. & really really afraid public embarrass awkward stat train station saw homeless laid ask food money away help homeless people chang wallet walk took wallet pull cash give kinda old perfect physics health & 0 & 1  \\ 
\hline
\end{tabular}
\label{example}
\end{table}

The models' accuracies have been calculated and are presented in Table \ref{table:nonlin}. Notably, the Logistic Regression model achieves the highest accuracy at 77.78\%.


\begin{table}[tb]
\label{result_trained}
\caption{Classification results of ML algorithms paired with BoW}
\begin{tblr}{|p{1.2cm}|p{1.2cm}|p{1.3cm}|p{1cm}|p{0.7cm}|p{0.7cm}|
}
\hline
\textbf{Features} & \textbf{ML} & \textbf{Accuracy,\%} & \textbf{Precision} & \textbf{Recall} & \textbf{F score} \\ \hline
BoW & SVM & 69.90 & 0.69 & 0.50 & 0.58\\ \hline
BoW & {Naive \\ Bayes} & 71.31 & 0.69 & 0.61 & 0.65 \\ \hline
\textbf{BoW} & \textbf{Logistic \\ Regression} & \textbf{77.78} & \textbf{0.80} & \textbf{0.78} & \textbf{0.79}\\ \hline
Word Embeddings & LSTM & 70.2 & 71.23 & 52.10 & 60.19 \\ \hline

\end{tblr}
\label{table:nonlin} 
\end{table}



We use metrics such as Accuracy, Precision, Recall, and $F_1$ score to evaluate the classification algorithms.  Precision estimates how many positively recognized samples are correct, while Recall estimates the fraction of positive samples correctly identified. The higher the F1 score, the closer both values are. 


\textbf{Precision}. Correct positive predictions relative to total positive predictions \cite{f1scoreAcc}:
\newline
\[Precision = \frac{TP}{TP + FP}\]

\textbf{Recall}. Correct positive predictions relative to total actual positives \cite{f1scoreAcc}: \newline
\[Recall = \frac{TP}{TP + FN}\]

$F_1$ score is \cite{f1scoreAcc}:\newline
\[F_1 = 2 * \frac{Precision * Recall}{Precision + Recall} \]

\textbf{Accuracy}. Correct predictions to the total prediction \cite{f1scoreAcc}:\newline
\[Accuracy =  \frac{TP +TN}{TP +TN+FP+FN} \]

 Our results are closely aligned with the findings of the creators of the Dreaddit dataset \cite{Turcan2019}. Their best-performing model achieved an F1-score of 0.798 using Logistic Regression with domain-specific Word2Vec and additional features. Using the Bag of Words approach paired with Logistic Regression, our model achieved a comparable F1 score of 0.79 on the same dataset. 

\subsubsection{Reddit Academic Dataset}
Let us use the human-annotated dataset discussed earlier to evaluate the performance of the proposed approach.

\begin{figure}
    \centering
    \includegraphics[width=0.5\textwidth]{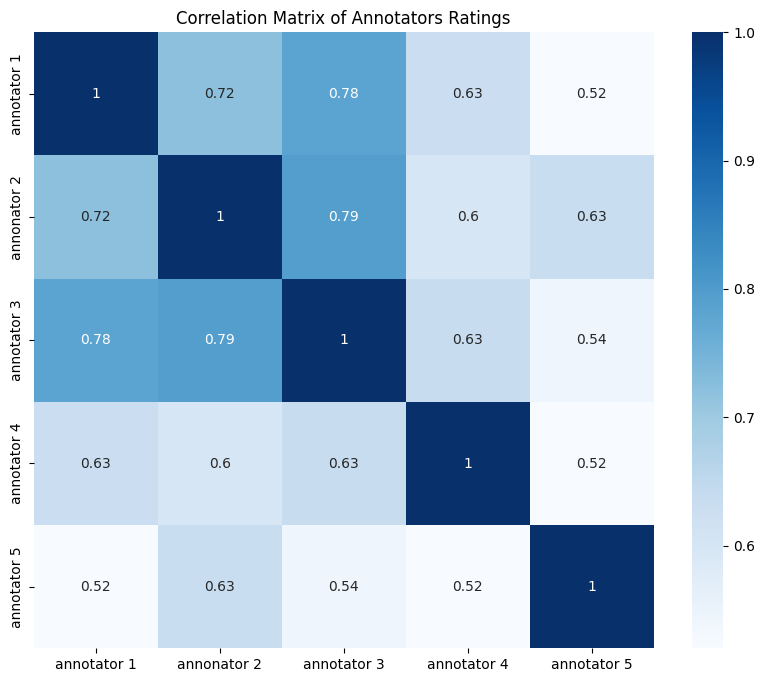}
    \caption{The heat map showing correlation of annotators' rating.}
    \label{agreement_heatmap}
\end{figure}

In the annotation experiment, we classified 100 post texts using the proposed model (BoW + Logistic Regression). The experiment involved five participants, including one psychologist and four outliers. The psychologist's labels were given higher importance due to their expertise, indicated by a weight of 2. The calculated accuracy of 72\% reflects the model's performance in correctly classifying the texts based on the input from all participants, with a higher emphasis on the psychologist's judgments. Figure \ref{agreement_heatmap} shows the heat map with correlations of ratings provided by annotators. We can see that annotator 3 correlates highly with annotators 1 and 2.

The calculated agreement on labeled data is $k \approx 0.13$, using Fleiss’s Kappa inter-annotator agreement \cite{annagr}. According to the interpretation of Fleiss's Kappa, the $k$ value shows \textit{slight agreement} between annotators.

\begin{figure}[t]
\centerline{\includegraphics[width=9cm,scale=0.3]{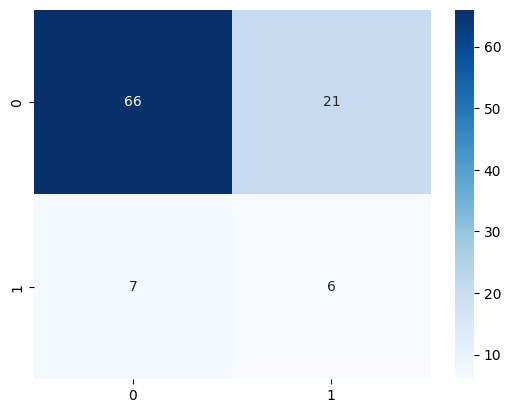}}
\caption{Confusion matrix.}
\label{confusion_matrix}
\end{figure}

The model's performance is assessed using a confusion matrix, which provides insights into the model's ability to classify texts accurately. The confusion matrix for our experiment is as follows in Figure \ref{confusion_matrix}.




\begin{table*}[!htb]
\centering
\caption{Mean, median, and standard deviation of upvotes of posts and comments that were classified as "stressed"/"not stressed"}
\begin{tblr}{|l|p{1.9cm}|p{1.9cm}|p{2cm}|p{2cm}|p{2cm}|p{2cm}|}
\hline
\textbf{Dataset}&\textbf{Stressed upvote mean}&\textbf{Stressed upvote median}&\textbf{Stressed upvote std}&\textbf{Not stressed upvote mean}&\textbf{Not stressed upvote median}&\textbf{Not stressed upvote std}
\\
\hline
Bachelor students& 39.5&2.0&255.1&27.0&2.0 &163.6 \\
\hline
Graduate students&26.7&4.0&97.6&19.9&3.0&71.4 \\
\hline
Ph.D. students&20.4&3.0&67.6&13.8&2.0&50.7 \\
\hline
Professors&35.6&7.0&120.8&25.7&6.0&84.2 \\
\hline
Human-annotated dataset&41.1&9.0&70.5&12.5&2.5&26.0 \\
\hline
\end{tblr}
\label{positivity_bias}
\end{table*}

\begin{figure}
    \centering
    \includegraphics[width=0.48\textwidth, scale=0.3]{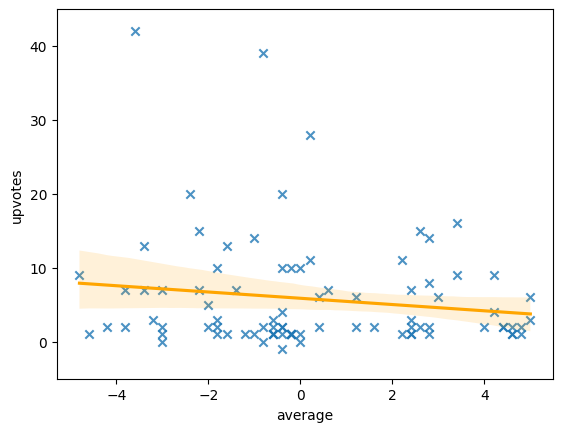}
    \caption{Regression plot of upvotes and average of annotators rating.}
    \label{regplot}
\end{figure}

Our results do not corroborate the findings of \cite{diffusion} on the influence of the sentiment on information diffusion. According to their research, positive messages reach a larger audience, meaning that individuals are more willing to repost positive information, a phenomenon known as positive bias. Figure \ref{regplot} shows upvotes of posts and comments as a function of the average stress score assigned by the annotators. The graph shows that most viral posts are neutral or slightly stressed, rather than excessively not stressed.

Table \ref{positivity_bias} shows us the mean and median upvotes of posts and comments of different academic levels in the human-annotated dataset classified as \textit{stressed/not stressed}. By the phenomenon of positivity bias, it is said that people tend to like positive views on life (not stressful) rather than negative (stressful) \cite{diffusion}. It also implies that people like and share online posts that do not evoke negative emotions.  However, after analyzing our datasets, we see that this does not apply to our case. This can possibly be explained by the nature of the Reddit social network.

\subsection{Stress detection Results - Academic Communities in Reddit}

\begin{table*}[tb]

\caption{Stress Detection results in various academic levels. The mean stress level is 29\%} 
\centering
  \begin{tblr}{
    width = \linewidth, 
    colspec = {|c|p{2.4cm}|c|c|c|c|c|c|c|c|},
  }
\hline
\textbf{Level in Academia} & \textbf{Subreddit name }& \textbf{\# of posts} & \textbf{\# of comments} &\textbf{Total}  & \textbf{Stressed} & \textbf{Stressed, \%} & \textbf{Not stressed} & \textbf{Not stressed, \%} \\ \hline
Bachelor students &r/csMajors, r/EngineeringStudents &100 &18281& 18381& 5389 &29.3\%&12992  &70.7\% \\ \hline
Graduate students &r/GradSchool& 785 &44762&45547& 14156 & 31.1\%&31391  & 68.9\%\\ \hline
PhD students &r/PhD  & 386 &22406&22792 &5647&24.8\% &17145  & 75.2\%\\ \hline
Professors &r/Professors  & 313& 37235 &37548 &11455 &30.5\%&26093 &69.5\%  \\ \hline

\end{tblr}
\label{final} 
\end{table*}

For our experiment, we analyzed posts and comments in popular Reddit communities of professors, bachelors, graduates, and Ph.D. students.  The most successful model trained on the Dreaddit dataset utilizes the BoW representation in combination with Logistic Regression. This model achieved an accuracy rate of 77.78\% and $F_1$ score of 0.79. So, this model is used to make an experiment using the dataset that we parsed from Reddit.

Table \ref{final} presents the stress detection results in considered academic levels: professors, bachelor, graduate, and Ph.D. students.


\begin{figure}[tb]
    \centering
    \begin{subfigure}{0.49\linewidth}
        \includegraphics[width=\linewidth]{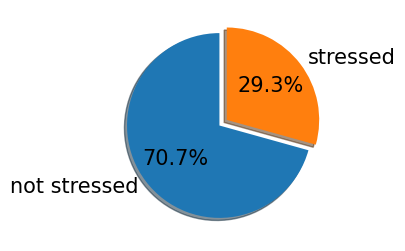}
        \caption{Bachelor students}
    \end{subfigure}
    \begin{subfigure}{0.49\linewidth}
        \includegraphics[width=\linewidth]{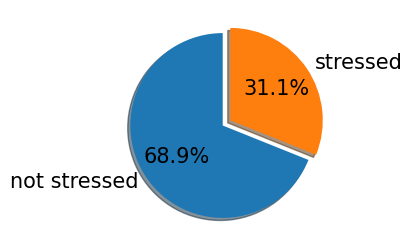}
        \caption{Graduate students}
    \end{subfigure}
    \begin{subfigure}{0.49\linewidth}
        \includegraphics[width=\linewidth]{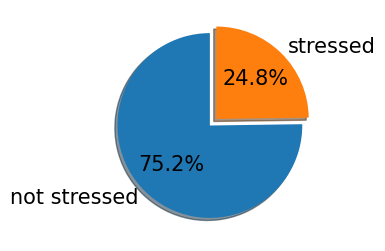}
        \caption{PhD students}
    \end{subfigure}
    \begin{subfigure}{0.49\linewidth}
        \includegraphics[width=\linewidth]{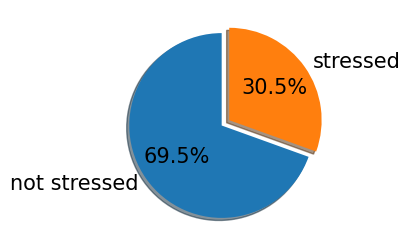}
        \caption{Professors}
    \end{subfigure}
    \caption{Pie chart representing the percentage ratio of stressed/not stressed in posts and comments of different academic levels.}
    \label{pie_chart}
\end{figure}


We used NRCLex \footnote{https://github.com/metalcorebear/NRCLex} Python library \cite{nrc1, nrc2} to retrieve negative emotional affect(anger, fear, sadness, disgust, surprise) from Reddit data that were classified as \textit{stressed}. NRCLex is based on the National Research Council Canada (NRC) affect lexicon and the NLTK library's WordNet synonym sets. NRCLex analyzes the text based on the words used and returns the frequency of emotional affect within the text. We filtered emotions with a value of 0.3 or higher to capture the most prevalent ones.

 
Table \ref{stress_example} represents some comments on Reddit from different levels of academia that were classified as \textit{stressed}. Along with academic level and text, emotion frequency was detected with NRCLex.

\begin{table}[bt]
\caption{Example of post/comments that were classified as "stressed"}
\label{stress_example}
    \centering
    \begin{adjustbox}{width=\linewidth}
    \begin{tabular}{|m{2cm}|m{5cm}|m{2cm}|}
    \hline
        \textbf{Academic Level} & \textbf{Text} & \textbf{Prevaling emotion} \\ 
        \hline
        Professors & My grandfather died the day before an exam. I attended the exam in mourning clothes. & sadness: 0.33\\
        \hline
        PhD & You don't want to hear this but it's worse in industry. They tell you this then just fire you :/ & fear: 1.0 \\
        \hline
        Graduates & I understand how you feel.  It is a very frustrating thing to be trying to tackle grad school and mental health at the same time! :( & anger: 0.33 \\
        \hline
        Bachelors &Is it too late to change if I wasted my freshman year (going into sophomore in the fall)? Always felt and still feel too intimidated to try these things.'& sadness: 0.5 \\
        \hline
    \end{tabular}
     \end{adjustbox}
\end{table}

\begin{table}[]
\caption{10 most frequently used words in posts/comments that were classified as "stressed" of different academia levels}
\begin{tabular}{|c|c|c|c|}
\hline
\textbf{Bachelors}       & \textbf{Graduates}         & \textbf{PhD} & \textbf{Professors} \\ 
\hline
'work', 1290 & 'work', 6762& 'phd', 2700&'student', 10841\\
'get', 1174 & 'school', 5220 & 'work', 2665 &'class', 5076 \\
'people', 1123 & 'get', 4871 & 'get', 2280& 'work', 3189 \\
'like', 1100 &  'student', 4835 & 'like', 2095 & 'one', 3017 \\
'class', 1036 & 'time', 4833 & 'time', 2030 & 'get', 2979  \\
'engine', 997& 'like', 4266 & 'people', 1520 & 'would', 2539 \\ 
'time', 879 & 'go', 4247 & 'go', 1472& 'time', 2487 \\
'go', 834 & 'grad', 4126 & 'think', 1438 & 'make', 2353 \\
'one', 806 & 'people', 3704 & 'know', 1337 & 'like', 2312 \\
'say', 794 & 'think', 3677 & 'one', 1248 & 'think', 2113 \\
\hline
\end{tabular}
\label{lexicon_table}
\end{table}
Table \ref{lexicon_table} shows a lexicon of more common terms among academic stressed data. The most common words at all academic levels are ``work'', ``time'', and ``get''. It is noticeable that graduate students and professors use the word ``student'' more frequently than ``people'' than bachelor and Ph.D. students. It is also interesting that undergraduate students are more likely to use the word ``say'' when others are more often using the word ``think''.

\begin{figure*}[hbtp]
    \centering
    \begin{subfigure}{0.5\textwidth}
        \includegraphics[width=\textwidth]{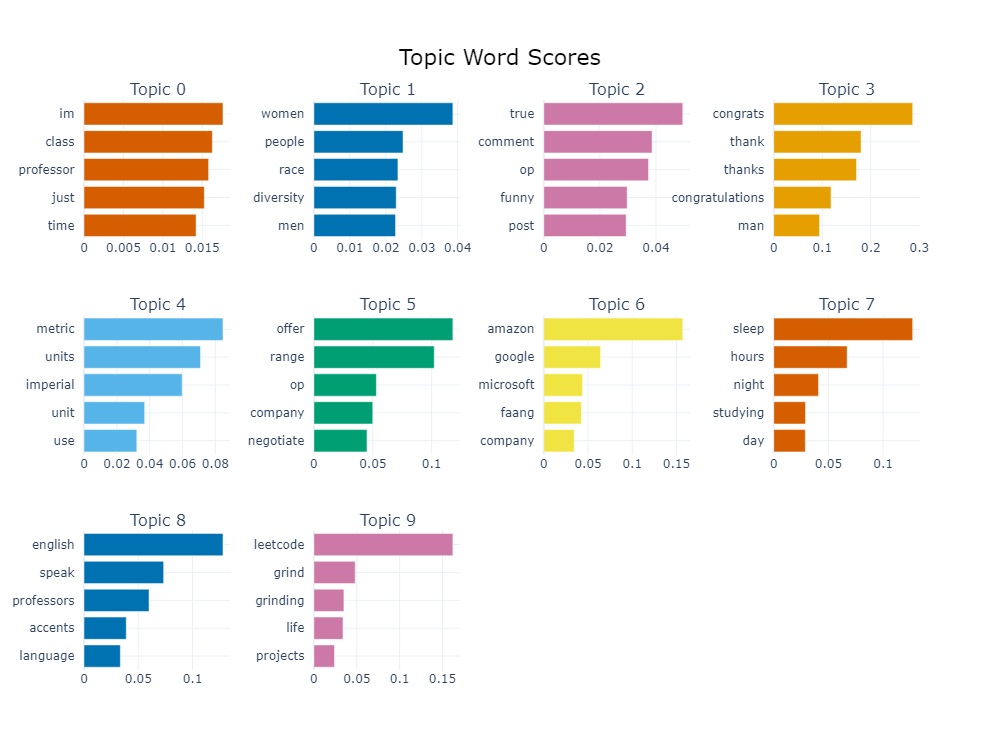}
        \caption{Bachelor students}
    \end{subfigure}%
    \begin{subfigure}{0.5\textwidth}
        \includegraphics[width=\textwidth]{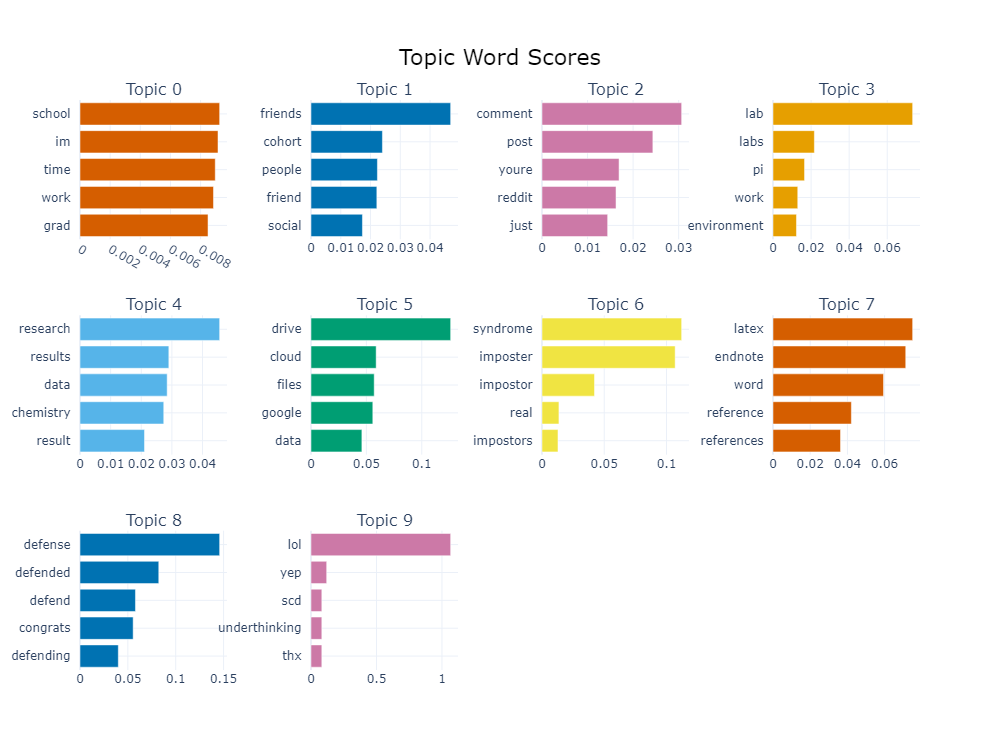}
        \caption{Graduate students}
    \end{subfigure}%
    \hfill
    \begin{subfigure}{0.5\textwidth}
        \includegraphics[width=\textwidth]{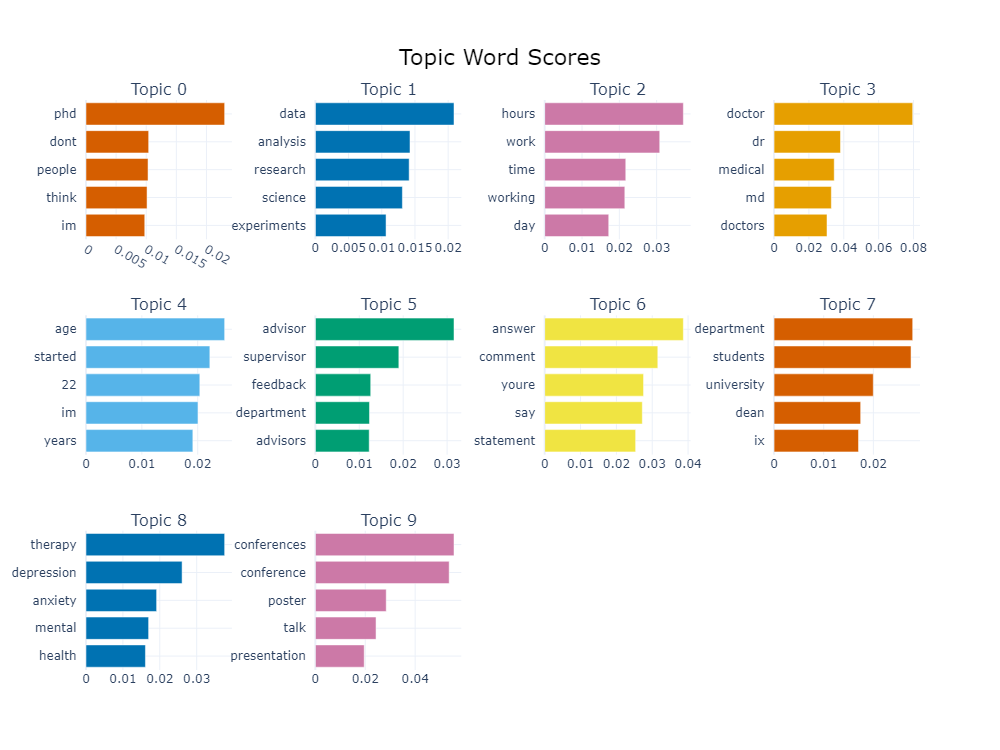}
        \caption{PhD students}
    \end{subfigure}%
    \begin{subfigure}{0.5\textwidth}
        \includegraphics[width=\textwidth]{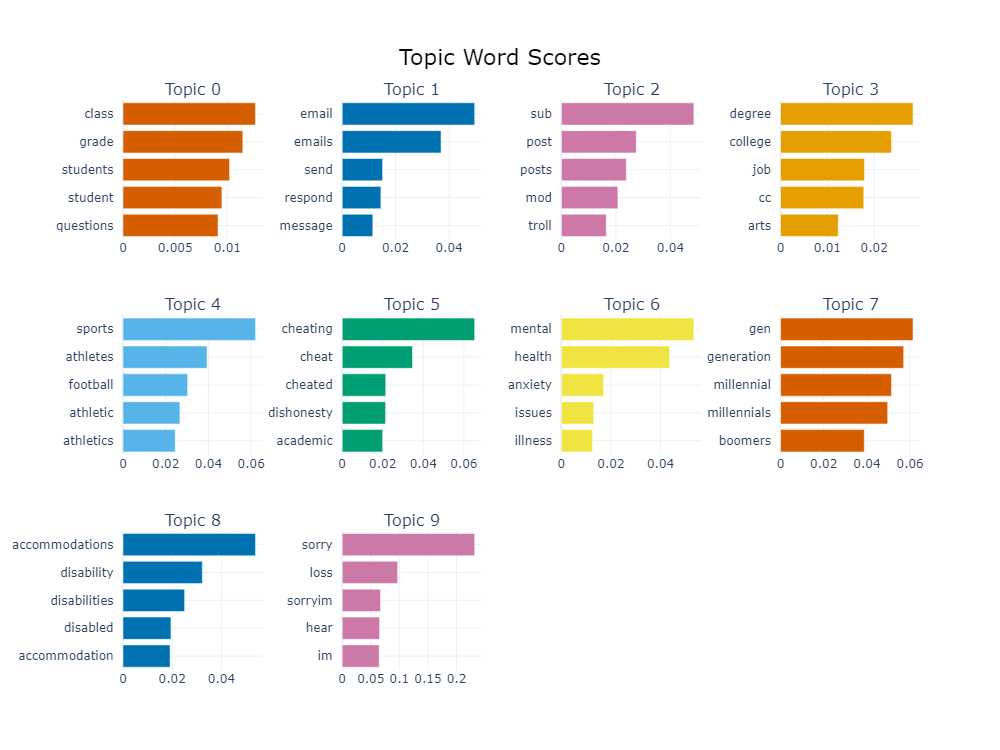}
        \caption{Professors}
    \end{subfigure}%
    \caption{Topics with top 5 words using BERT.}
    \label{topic_models}
\end{figure*}

Words alone don't tell us much about stressed posts and comments. To get an idea of the key factors contributing to academic stress, we performed topic modeling, and the results can be seen in Figure \ref{topic_models}. For identifying topics, we used the BERT language representation model\cite{bert_article}. BERT, which stands for Bidirectional Encoder Representations from Transformers, is a comprehensive tool for analyzing short texts with little context and is less time-consuming. We can see what kinds of words are frequently used together, and it gives us a little understanding of topics mostly discussed on Reddit posts and comments that can be considered as \textit{stressed}. The posts and comments about \textit{studying} and \textit{classes}, \textit{professors} and their \textit{English skills}, \textit{major IT companies }and \textit{internships}, and \textit{sleep }resonate the most with stress among bachelor students. \textit{Balancing jobs and education}, \textit{research}, and \textit{defense of a master's dissertation} probably are stressful for graduate students. \textit{Research and supervisors}, \textit{working}, \textit{mental health} and \textit{therapy}, and \textit{talking to big audiences at conferences }are topics mostly associated with stress among Ph.D. students. Everything connected with \textit{teaching} and \textit{students}, \textit{answering their questions}, \textit{grading them}, \textit{classes}, and \textit{responding to emails} is stressful for professors.

\begin{figure}[!htbp]
\centerline{\includegraphics[width=0.5\textwidth]{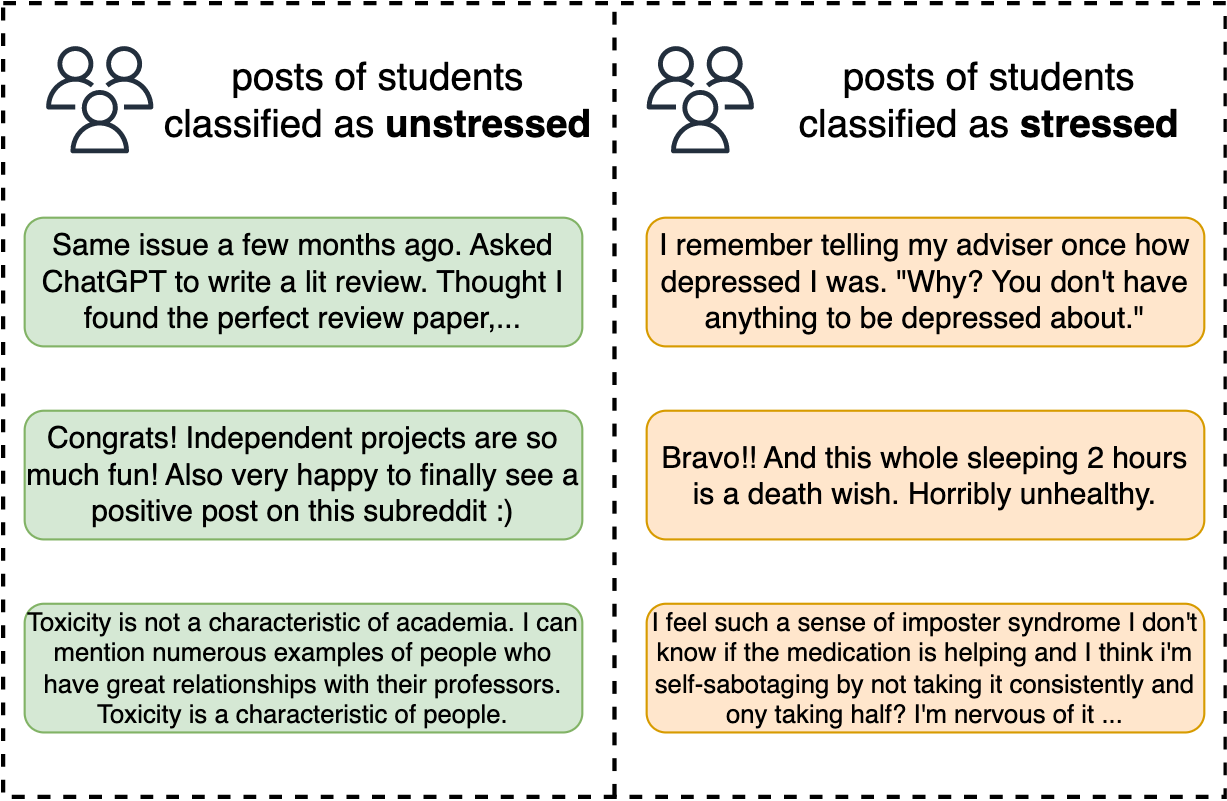}}
\caption{Examples of students' posts (or comments) classified as stressed and unstressed.}
\label{examples}
\end{figure}

Some sample students' posts (or comments) classified as stressed and unstressed are presented in Figure \ref{examples}.

\begin{figure*}[!htbp]
    \centering
    \includegraphics[width=0.8\textwidth]{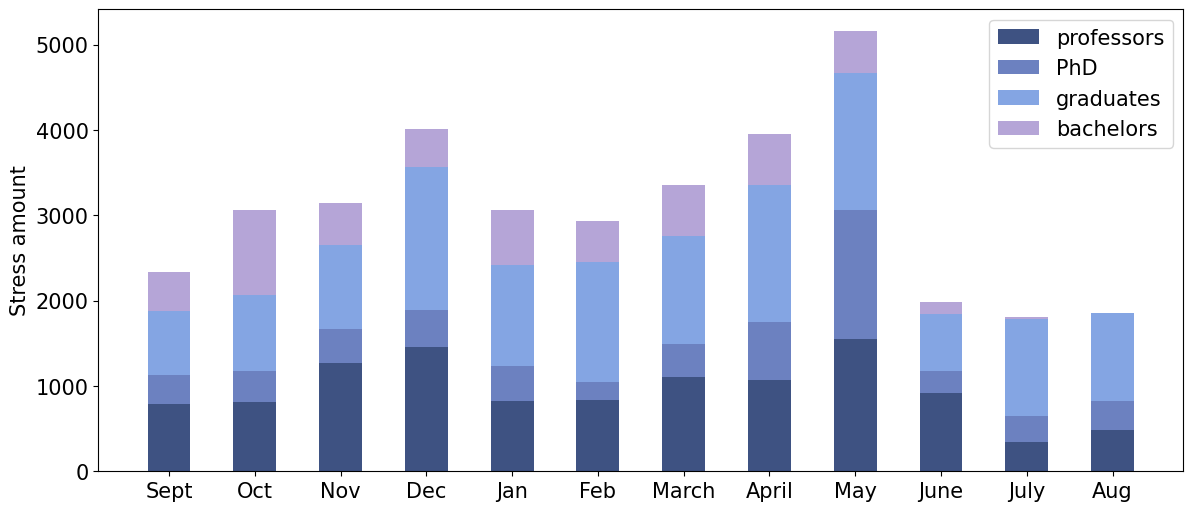}
    \caption{Distribution of posts/comments classified as "stressed" by months. The stacked bar charts illustrate the amount of stressed Reddit posts/comments from different levels of academia.}
    \label{stress_bymonth}
\end{figure*}

Figure \ref{stress_bymonth} shows the stress change among professors, bachelor, graduate, and Ph.D. students during the year (stress tracking). Posts/comments classified as stressed were categorized monthly, starting in September and ending in August, to match the academic year.  The illustrated bar charts show that professors and graduate students tend to be stressed throughout the year. Specifically, there are noticeable spikes in stress for all levels of academia, which coincides with months (December and May) when exams are usually conducted. Bachelor students' stress levels remain relatively stable throughout the academic year except in October, and stress levels are lower during summer. It may appear that bachelor students experience less stress compared to other academic levels, but this is mostly due to the smaller size of the parsed Reddit dataset used for analysis.

\begin{figure*}[!hbtp]
  \begin{subfigure}{0.5\textwidth}
    \includegraphics[width=\textwidth]{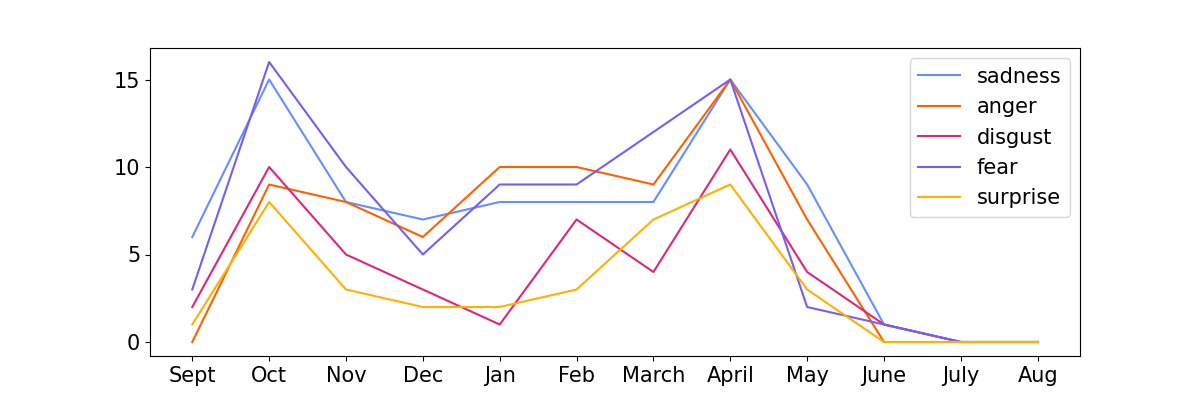}
    \caption{Bachelor students} 
  \end{subfigure}%
  \begin{subfigure}{0.5\textwidth}
    \includegraphics[width=\textwidth]{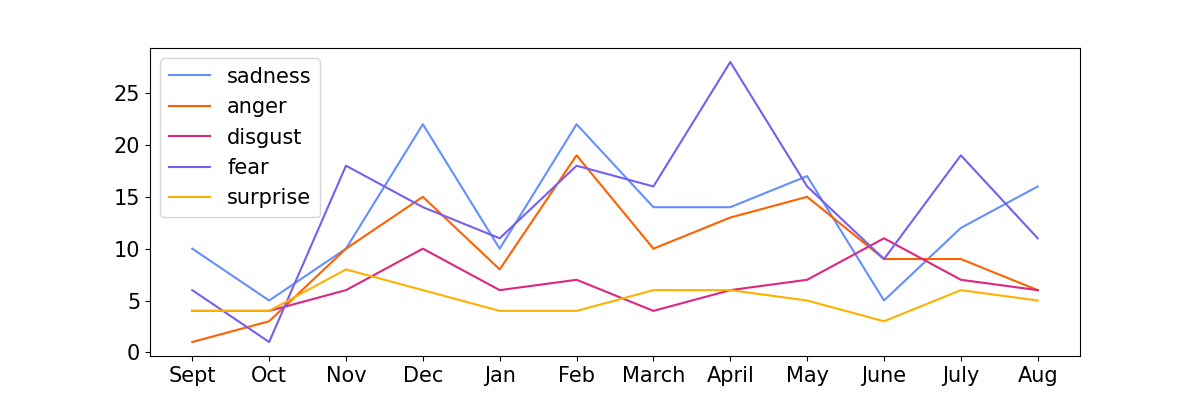}
    \caption{Graduate students} 
  \end{subfigure}%

  \begin{subfigure}{0.5\textwidth}
    \includegraphics[width=\textwidth]{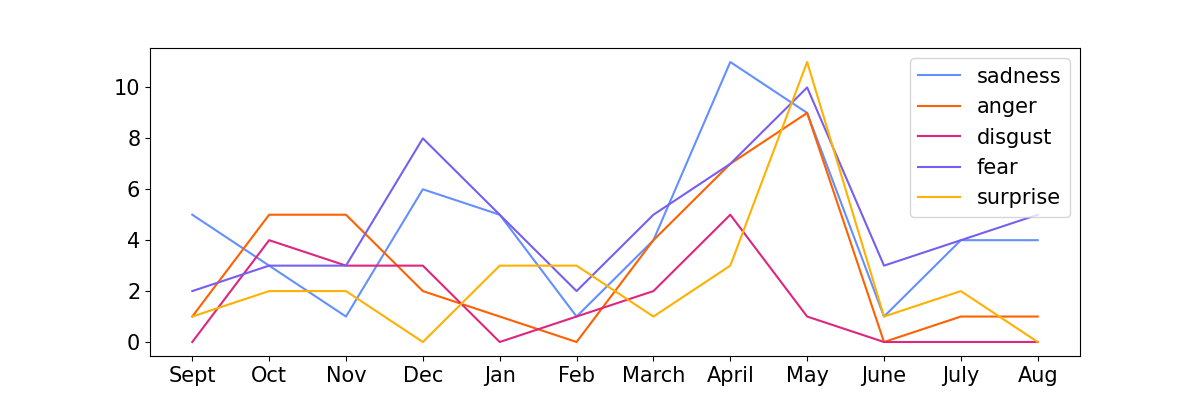}
    \caption{PhD students} 
  \end{subfigure}
    \begin{subfigure}{0.5\textwidth}
    \includegraphics[width=\textwidth]{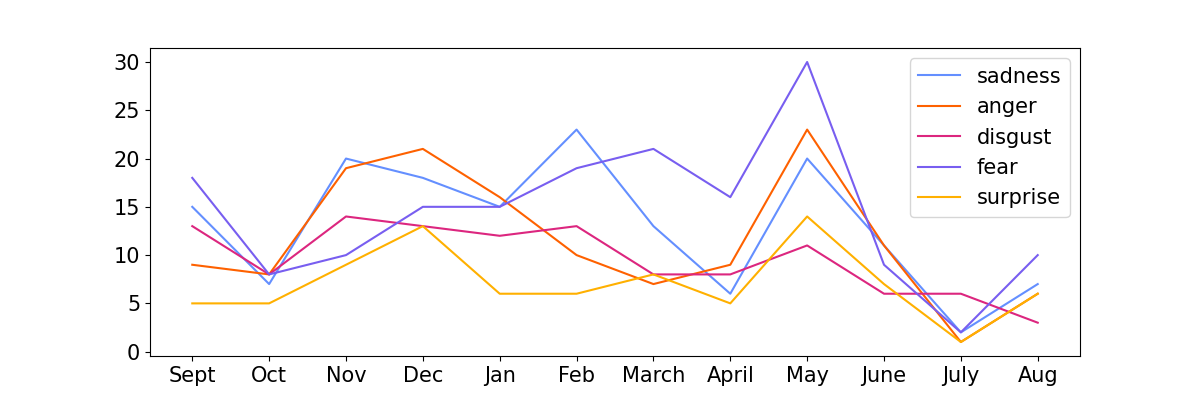}
    \caption{Professors} 
  \end{subfigure} 
     
\caption{Prevailing emotions in Reddit posts/comments classified as "stressed" by months in different levels of academia.} 
\label{emotion_bymonth}
\end{figure*}








\begin{figure*}[!htbp]
  \begin{subfigure}{0.5\textwidth}
  \centering
    \includegraphics[width=0.7\linewidth]{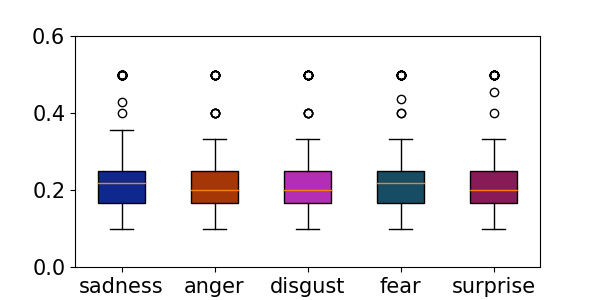}
    \caption{Bachelor students} 
  \end{subfigure}%
  \begin{subfigure}{0.5\textwidth}
  \centering
    \includegraphics[width=0.7\linewidth]{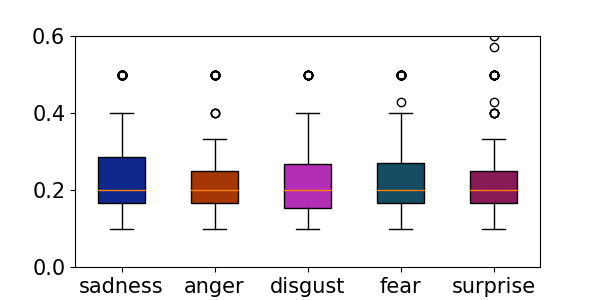}
    \caption{Graduate students} 
  \end{subfigure}%

  \begin{subfigure}{0.5\textwidth}
  \centering
    \includegraphics[width=0.7\linewidth]{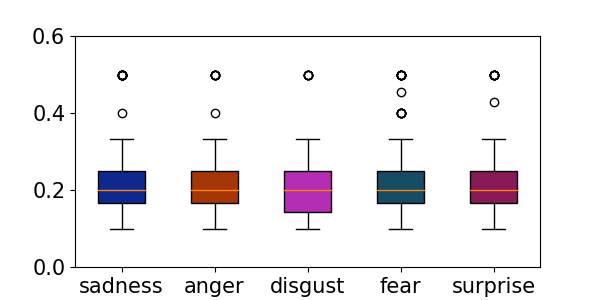}
    \caption{PhD students} 
  \end{subfigure}
    \begin{subfigure}{0.5\textwidth}
    \centering
    \includegraphics[width=0.7\linewidth]{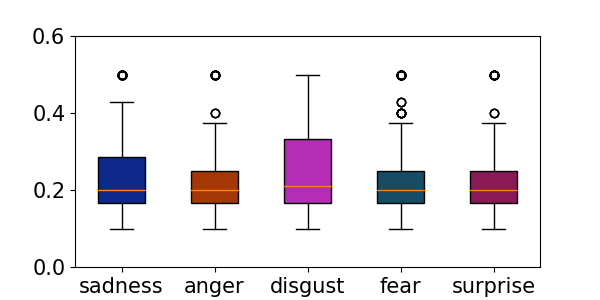}
    \caption{Professors} 
  \end{subfigure}

\caption{Summarising emotions in posts/comments classified as "stressed". The whisker plots show the distribution of detected emotion values in stressed Reddit posts/comments of different levels of academia.} 
\label{whisker_plot}
\end{figure*}

Figures \ref{emotion_bymonth} and \ref{whisker_plot} present line and whisker plots depicting the distribution of identified emotion ratings in stressed Reddit posts/comments from various academic levels. Line plots represent how emotions change throughout the academic year. A clear pattern for all academic levels is a low level of negative emotions in the summer months, which then rises from September until the end of the academic year. As we can see prevailing emotions of stressful posts across all academic levels are sadness and fear. The whisker plot illustrates how emotions are overall distributed and the outliers that appear there.


\section{Discussion}

Let us consider how the current research findings compare to previous studies. Mental stress has become a significant issue among young people, affecting various aspects of their lives. 

With nearly 55 million daily active users and millions of posts publicly broadcasted to a large audience, Reddit is one of the most popular social networking sites \cite{stress-dep2}. Several works employ machine-learning algorithms to detect stress \cite{disc1}, \cite{disc2}. Some works focused on the identification of the presence of depression in Reddit texts using ML and NLP techniques \cite{stress-dep2}, \cite{depr}, Reddit Comment Toxicity Score Prediction using BERT \cite{toxic}.

Our findings support previous research that found stress factors in students: academic, teaching, and learning-related stressors \cite{stressFactors_nbci}. The same topics were detected in our research and are shown in Figure \ref{topic_models}. In contrast, our results presented in Table \ref{positivity_bias} do not support the phenomenon of positive bias discussed in \cite{diffusion}, according to which people tend to share and like news and posts on the Internet that do not evoke negative emotions.

We challenge the findings of \cite{stressFactors_nbci}, asserting that lower-grade students experience more stress than sixth-year medical students. Our research findings indicate that students in higher grades, specifically those pursuing Master's degrees (the same age as medical students in the 5th–6th grades), demonstrate the highest stress levels. This is likely attributed to increased responsibilities, the demand for studying complex topics, completing a Master's dissertation, and the challenges of combining academic work with other commitments.

As we see, previously, similar research had been conducted, but on much broader topics such as mental health, depression, disorders, etc. We narrowed our focus and studied the stress levels in academia and the associated emotions and topics. We also analyze the stress levels of various academic representatives and do not rely on self-reporting (through questionnaires). Studies investigating such broad concepts as stress based on questionnaires are limited \cite{covidstress}.

 \section{Conclusion}

This research uses ML and NLP techniques to recognize and evaluate stress-related posts and comments in Reddit's academic communities. 

Automatic stress detection is crucial in mitigating stress risks and improving mental health outcomes. By enabling early detection, addressing hesitancy to seek help,  and providing a preliminary assessment, such systems can revolutionize stress management and positively impact individuals' well-being. Stress detection from textual data in social media platforms presents a promising solution, considering its effectiveness and ease of access to data.


We achieved the highest predictive performance on the Dreaddit dataset with Bag of Words and Logistic regression, with an accuracy of 77.78\% and an F1 score of 0.79. To validate our model's applicability in academia, we conducted a supplementary experiment by manually annotating 100 posts from academic subreddits, achieving a 72\% accuracy rate.

Our findings suggest that the overall stress level in academic texts is 29\%, with the key emotions being sadness and fear. Moreover, the key factors contributing to this stress are:
\begin{itemize}
\item 
For bachelor students, these include studying classes, professors and their English skills, major IT companies, internships, and sleep.
\item 
For graduate students - balancing job and education, research, and defense of a master's dissertation.  
\item 
For PhD students, it is research and their supervisors, mental health and therapy, and talking to big audiences at conferences.
\item 
For professors -  students, and teachers, responding to questions and emails.
\end{itemize}

Our findings can assist universities in building better-targeted interventions and support systems by identifying and quantifying stress levels at different academic levels early. This tailored approach to managing academic stress-related issues can improve students' and faculty's mental health and well-being. Administrators and educators can use the current study's findings to adopt policies and practices that reduce stress and foster a better learning and working environment.

The study has some limitations.  Estimates of academic stress levels were based solely on one social network, Reddit. The other limitation is the relatively small dataset size. These are issues that need to be addressed in future work. We also plan to experiment with other Machine Learning techniques (Bert, Random Forest, Decision Tree, XGBoost) and Computing with Words technique \cite{cw2, cw1} to increase the accuracy.

The majority of approaches for stress identification focus on physical aspects only or require a psychological test, which can lead to inaccurate results. Stress can manifest through various modalities, including text, speech, and physiological signals. Integrating these diverse data sources to improve detection accuracy is a complex but important challenge we plan to explore.

\bibliography{exp-bib}

\begin{IEEEbiography}[{\includegraphics[width=1in,height=1.25in,clip,keepaspectratio]{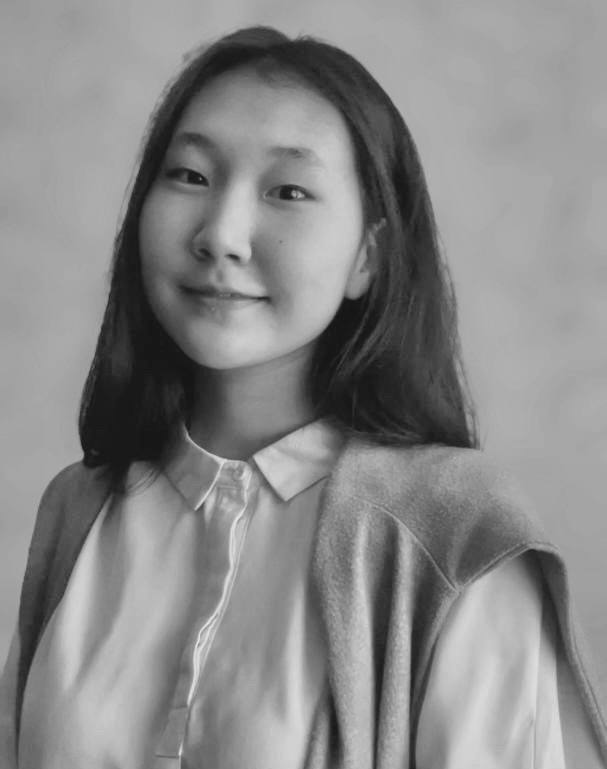}}]{Nazzere Oryngozha} received the B.S. degree in systems of information security from
the  International Information Technology University, Almaty, Kazakhstan, in 2022.  Since 2022, he has been a tutor with the
Cybersecurity Department, at the same university. She is currently pursuing M. S. degree in software engineering at the School of Information Technology and Engineering, Kazakh-British Technical University, Almaty, Kazakhstan, and works as a backend developer (Python).


\end{IEEEbiography}

\begin{IEEEbiography}[{\includegraphics[width=1in,height=1.25in,clip,keepaspectratio]{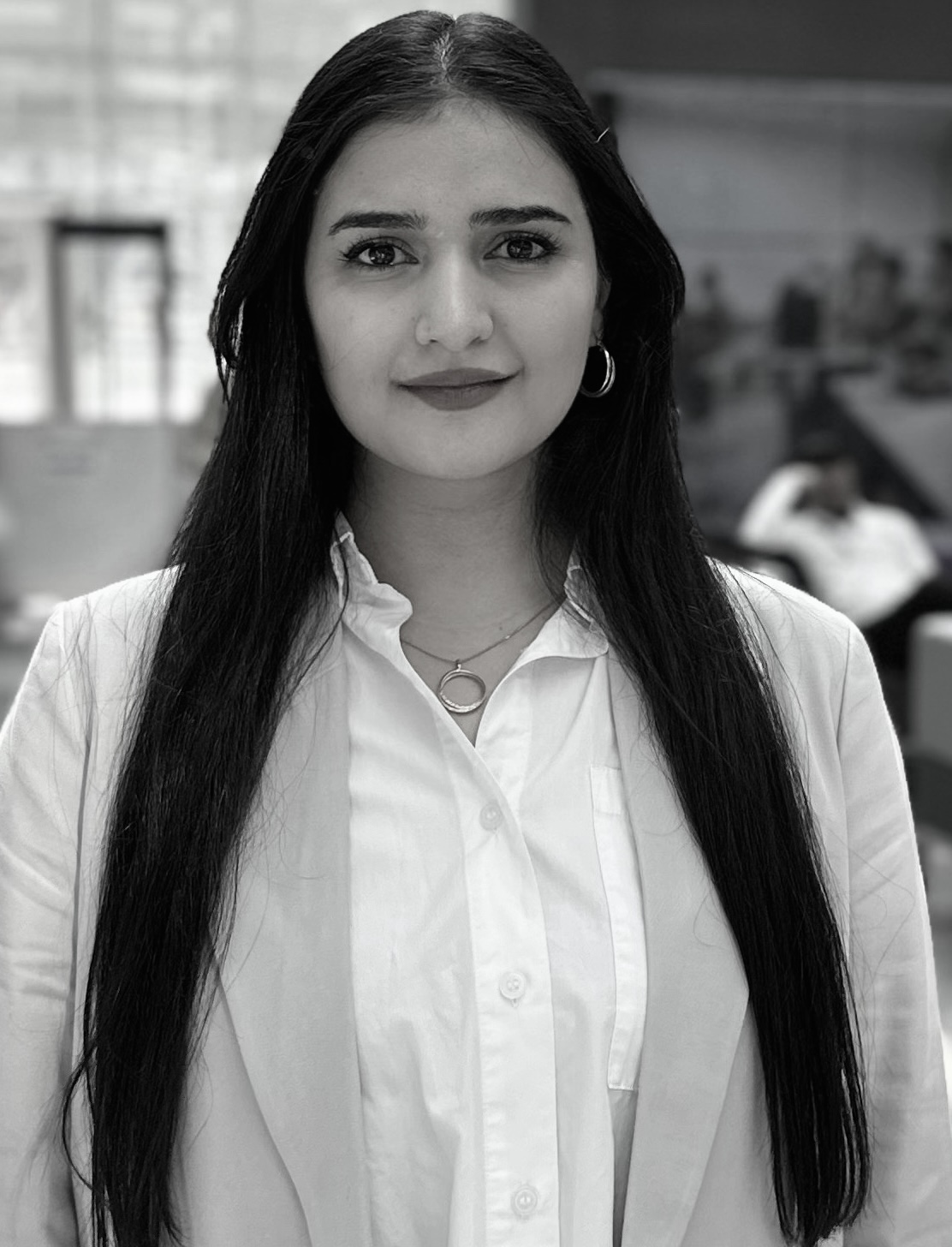}}]{Pakizar Shamoi}  received the B.S. and M.S. degrees in information systems from the Kazakh-British Technical University, Almaty, Kazakhstan in 2011 and 2013, and the Ph.D. degree in engineering from Mie University, Tsu, Japan, in 2019. In her academic journey, she has held various teaching and research positions at Kazakh-British Technical University, where she has been serving as a professor in the School of Information Technology and Engineering since August 2020. Her commitment to education and contributions to the field have been recognized with the prestigious Best University Teacher Award in Kazakhstan in 2022. She is the author of 1 book and more than 28 scientific publications. Awards for the best paper at conferences were received four times. Her research interests include artificial intelligence and machine learning, focusing on fuzzy sets and logic, soft computing, representing and processing colors in computer systems, natural language processing, computational aesthetics, and human-friendly computing and systems. She took part in the organization and worked in the org. committee (as head of the session and responsible for special sessions) of several international conferences - IFSA-SCIS 2017, Otsu, Japan; SCIS-ISIS 2022, Mie, Japan; EUSPN 2023, Almaty, Kazakhstan. She served as a reviewer at several international conferences, including IEEE:
SIST 2023, SMC 2022, SCIS-ISIS 2022, SMC 2020, ICIEV-IVPR 2019, ICIEV-IVPR 2018.

\end{IEEEbiography}

\begin{IEEEbiography}[{\includegraphics[width=1in,height=1.25in,clip,keepaspectratio]{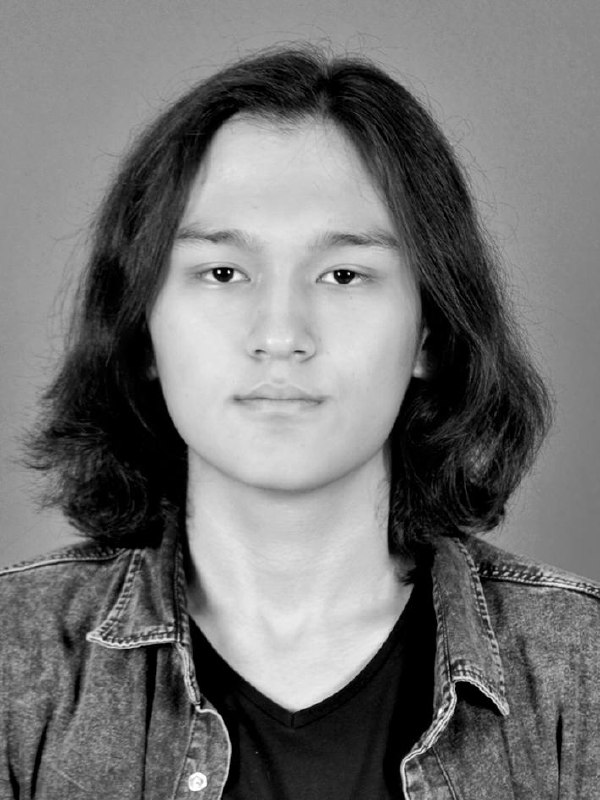}}]{Ayan Igali} is currently a 3rd-year bachelor student at the School of Information Technology and Engineering, Kazakh-British Technical University, Almaty, Kazakhstan, his major is information systems. His research interests include Machine Learning, NLP, emotion detection, and human-friendly systems.
\end{IEEEbiography}

\EOD
\end{document}